\newcommand{\hide}[1]{}

\documentclass[preprint,review]{elsarticle}
\usepackage{rawfonts}

\usepackage{amsmath,amsfonts,amssymb,amsthm}
\usepackage{graphicx}
\usepackage{subfig}
\usepackage{psfrag}
\usepackage{xspace}
\usepackage{url}
\usepackage{bm}
\usepackage{latexsym}
\usepackage{epsfig}
\usepackage{makeidx}
\usepackage{enumerate}
\usepackage{booktabs}
\usepackage{multirow}
\usepackage{dsfont}
\usepackage[normalem]{ulem}
\usepackage[dvipsnames]{xcolor}
\usepackage{endnotes}
\usepackage{scalerel}
\usepackage{verbatim}
\usepackage{fancyvrb}
\usepackage{xifthen}
\usepackage{etoolbox}

\usepackage{tikz}
\usetikzlibrary{arrows}
\usetikzlibrary{shapes.geometric}
\usetikzlibrary{fit}

\usepackage{tabularx}

\usepackage{listings}
\usepackage[textsize=scriptsize,textwidth=1cm]{todonotes}

\usepackage[boxed,ruled,vlined,linesnumbered]{algorithm2e}
\SetAlCapFnt{\normalsize}
\SetAlgoInsideSkip{smallskip}
\SetAlFnt{\small}
\SetCommentSty{textit}
\SetInd{0.4em}{0.4em}

\newcommand{\papertitle}{Efficient Multi-agent Epistemic Planning: \\Teaching Planners About Nested Belief}

\DeclareSymbolFont{symbolsC}{U}{txsyc}{m}{n}
\SetSymbolFont{symbolsC}{bold}{U}{txsyc}{bx}{n}
\def\re@DeclareMathSymbol#1#2#3#4{%
    \let#1=\undefined
    \DeclareMathSymbol{#1}{#2}{#3}{#4}}

\re@DeclareMathSymbol{\Diamondblack}{\mathord}{symbolsC}{95}
\re@DeclareMathSymbol{\Diamond}{\mathord}{symbolsC}{94}

\newcommand{\towrite}[1]{\vspace{1mm}\hspace{3mm} $\langle \langle$ #1 $\rangle \rangle$\vspace{1mm}}

\newcommand{\citeasnoun}[1]{\citeauthor{#1} \cite{#1}}

\def\myisdef{\hbox{~$\stackrel{\text{def}}{=}$~}}

\newcommand{\la}{\langle}
\newcommand{\ra}{\rangle}
\newcommand{\set}[1]{\lbrace #1 \rbrace}

\renewcommand{\L}{\mathcal{L}}
\renewcommand{\P}{\mathcal{P}}
\newcommand{\Preg}{\mathcal{P}_{reg}}
\newcommand{\Pak}{\mathcal{P}_{AK}}

\newcommand{\A}{\mathcal{A}}
\newcommand{\D}{\mathcal{D}}
\newcommand{\sub}{_}
\newcommand{\agents}{Ag}
\newcommand{\allrmls}{\mathcal{L}^{\agents,d}_{RML}}

\newcommand{\fluents}{F}

\newcommand{\init}{\mathcal{I}}
\newcommand{\goal}{\mathcal{G}}
\newcommand{\operators}{O}

\newcommand{\pre}{Pre}
\newcommand{\eff}{\mathit{eff}}
\newcommand{\Eff}{\mathit{Eff}}
\newcommand{\poseffs}[1]{\eff^+_{#1}}
\newcommand{\negeffs}[1]{\eff^-_{#1}}
\newcommand{\cond}{\mathcal{C}}
\newcommand{\lit}{l}

\newcommand{\opseq}{\vec{o}}

\newcommand{\closure}{Cl}

\newcommand{\allworlds}{\mathcal{W}}

\newcommand{\progr}[2]{\mathit{Progress}\left(#1,#2\right)}
\newcommand{\entails}[1]{\models_{#1}}

\newcommand{\pekbentails}{\entails{PEKB}}

\newcommand{\optarg}[1]{\ifthenelse{\isempty{#1}}{}{(#1)}}
\newcommand{\optfirstarg}[2]{\ifthenelse{\isempty{#1}}{}{(#1,#2)}}

\newcommand{\close}[1]{\mathord{\uparrow}#1}
\newcommand{\dclose}[1]{\mathord{\downarrow}#1}

\newcommand{\update}{\mathbin{\raisebox{0.2mm}{\scaleobj{0.75}{\Diamond}}}}
\newcommand{\erase}{\mathbin{\raisebox{0.2mm}{\scaleobj{0.75}{\Diamondblack}}}}
\newcommand{\conjoin}{\mathbin{\sqcup}}
\newcommand{\disjoin}{\mathbin{\sqcap}}

\newcommand{\negkb}[1]{{\overline{#1}}}

\newcommand{\kbstring}{PEKB}
\newcommand{\kb}{\kbstring\xspace}
\newcommand{\kbs}{\kbstring s\xspace}

\newcommand{\literala}{\varphi}
\newcommand{\literalb}{\psi}

\def\schi{\chi_1 \land \ldots \land \chi_n}

\newcommand{\belief}[1]{\square_{#1}}
\newcommand{\possible}[1]{\Diamond_{#1}}
\newcommand{\invertedbelief}[1]{\mathcal{B}_{#1}}
\newcommand{\ninvertedbelief}[1]{\mathcal{N}_{#1}}

\newcommand{\actingsym}{\star}
\newcommand{\acting}{\belief{\actingsym}}

\newcommand{\condeff}[2]{(#1 \rightarrow #2)}
\newcommand{\condp}{\cond^{+}}
\newcommand{\condn}{\cond^{-}}

\newcommand{\macond}[2]{\mu^{#1}_{#2}}

\newcommand{\kd}{KD$_n$\xspace}
\newcommand{\kdff}{KD45$_n$\xspace}

\newcommand{\fullepplan}{\textsc{MEP}\xspace}
\newcommand{\epplan}{RP{-}\fullepplan}

\newcounter{definition}
\newenvironment{definition}[1]{
\refstepcounter{definition}
\noindent\begin{minipage}{\linewidth}
\vspace{2mm}
\noindent\hrulefill

\textbf{Definition \arabic{definition}.} #1

\vspace{-2.4mm}
\noindent\hrulefill
\vspace{1mm}
\end{minipage}
}
{

\vspace{-1.6mm}
\noindent\hrulefill
\vspace{1.6mm}}

\theoremstyle{definition}
\newtheorem{theorem}{Theorem}
\newtheorem{lemma}{Lemma}
\newtheorem{example}{Example}
\newtheorem{corollary}{Corollary}

\newcommand{\mm}[1]{\hspace{#1mm}}
\newcommand{\st}{\mm{1}|\mm{1}}

\newcommand{\tm}[1]{\vspace{0.5em}\todo[author=Tim,inline,color=green]{#1}\vspace{0.5em}}
\newcommand{\pf}[1]{\vspace{0.5em}\todo[author=Paolo,inline,color=red!30!white]{#1}\vspace{0.5em}}

\def\bbb{\belief{}\belief{}\belief{}p}
\def\bbp{\belief{}\belief{}\possible{}p}
\def\bpb{\belief{}\possible{}\belief{}p}
\def\pbb{\possible{}\belief{}\belief{}p}
\def\bpp{\belief{}\possible{}\possible{}p}
\def\pbp{\possible{}\belief{}\possible{}p}
\def\ppb{\possible{}\possible{}\belief{}p}
\def\ppp{\possible{}\possible{}\possible{}p}

\newenvironment{pdkbddl}
{\vspace{1em}\VerbatimEnvironment\begin{Verbatim}[frame=single,label=PDKBDDL,labelposition=topline]}
{\end{Verbatim}}

\newenvironment{labeledpdkbddl}[1][\unskip]
{\vspace{1em}\VerbatimEnvironment\begin{Verbatim}[frame=single,label=#1,labelposition=topline]}
{\end{Verbatim}}

\newenvironment{epddl}
{\vspace{1em}\VerbatimEnvironment\begin{Verbatim}[frame=single,label=EPDDL,labelposition=topline]}
{\end{Verbatim}}

\newcommand{\notecolour}{black!10!white}

\newcommand{\revision}[5]
{\vspace{0.5em}
    
    \ifstrequal{#1}{Done}{\renewcommand{\notecolour}{black!10!white}}{\renewcommand{\notecolour}{red!10!white}}

 \todo[inline,color=\notecolour]{

    \noindent\textbf{Status}: #1 \\
    \noindent\textbf{Assigned}: #2 \\
    \noindent\textbf{Estimate}: #3 \\
    \vspace{1em}\noindent\textbf{Notes}: #4 \\
    \vspace{1em}\noindent\textbf{Description}: #5 \\
    \vspace{1em}
 }
\vspace{0.5em}}

\renewcommand{\revision}[5]{{\color{BrickRed}\noindent #5} \\ \noindent\rule{\textwidth}{1pt}}

\begin{document}

\begin{frontmatter}

\title{\papertitle}

\author[1]{Christian Muise\corref{cor1}}
\ead{christian.muise@queensu.ca}
\address[1]{School of Computing \\ Queen's University, Kingston, Canada}

\author[2]{Vaishak Belle}
\ead{vaishak@ed.ac.uk}
\address[2]{University of Edinburgh, UK \& \\ Alan Turing Institute, UK}

\author[3]{Paolo Felli}
\ead{pfelli@unibz.it}
\address[3]{Faculty of Computer Science \\ University of Bozen-Bolzano, Bolzano, Italy}

\author[4]{Sheila McIlraith}
\ead{sheila@cs.toronto.edu}
\address[4]{Department of Computer Science \\ University of Toronto, Toronto, Canada}

\author[5]{Tim Miller}
\ead{tmiller@unimelb.edu.au}
\address[5]{School of Computing and Information Systems \\ University of Melbourne, Melbourne, Australia}

\author[5]{Adrian R. Pearce}
\ead{adrianrp@unimelb.edu.au}

\author[5]{Liz Sonenberg}
\ead{l.sonenberg@unimelb.edu.au}

\cortext[cor1]{Corresponding author}

\begin{abstract}

Many AI applications involve the interaction of multiple autonomous agents, requiring those agents to reason about their own beliefs, as well as those of other agents. However, planning involving nested beliefs is known to be computationally challenging. In this work, we address the task of synthesizing plans that necessitate reasoning about the beliefs of other agents. We plan from the perspective of a single agent with the potential for goals and actions that involve nested beliefs, non-homogeneous agents, co-present observations, and the ability for one agent to reason \emph{as if} it were another. We formally characterize our notion of planning with nested belief, and subsequently demonstrate how to automatically convert such problems into problems that appeal to classical planning technology for solving efficiently. Our approach represents an important step towards applying the well-established field of automated planning to the challenging task of planning involving nested beliefs of multiple agents.

\end{abstract}


\end{frontmatter}

\section{Introduction}
\label{sec:intro}

AI applications increasingly involve the interaction of multiple agents -- be they intelligent user interfaces that interact with human users, gaming systems, or multiple autonomous robots interacting together in a factory setting.  In the absence of prescribed coordination, it is often necessary for individual agents to synthesize their own plans, taking into account not only their own capabilities and beliefs about the world but also their beliefs about other agents, including what each of the agents will come to believe as the consequence of the actions of others.
To illustrate, consider the scenario where Larry and Moe meet on a regular basis at the local diner to swap the latest gossip. Larry has come to know that Nancy (Larry's daughter) has just received a major promotion in her job, but unbeknownst to him, Moe has already learned this bit of information through the grapevine. Before they speak, both believe Nancy is getting a promotion, Larry believes Moe is unaware of this (and consequently wishes to share the news), and Moe assumes Larry must already be aware of the promotion but is unaware of Moe's own knowledge of the situation. Very quickly we can see how the nesting of (potentially incorrect) belief can be a complicated and interesting setting to model.

\label{page:perspective-motivation}
In this paper, we examine the problem of synthesizing plans in such settings. In particular, given a finite set of agents, each with: (1) (possibly incomplete and incorrect) beliefs about the world and about the beliefs of other agents; and (2) differing capabilities including the ability to perform actions whose outcomes are unknown to other agents;  we are interested in synthesizing a plan to achieve a goal condition. Planning is at the belief level and as such, while we consider the execution of actions that can change the state of the world (ontic actions) as well as an agent's state of knowledge or belief (epistemic or more accurately doxastic actions, including communication actions), all outcomes are with respect to belief.  Further, those beliefs respect the \kd and \kdff axioms of epistemic logic \cite{Fagin}. Finally, we take a \emph{perspectival view}, planning from the viewpoint of a single agent. While the planning agent presumes control of all the other agents actions (i.e., we do not model the possibility of other agents acting in an uncertain and unobserved manner), there is \textit{not} presumption of complete knowledge over the nested belief of other agents. We contrast this with traditional multi-agent planning which generates a coordinated plan to be executed by multiple agents under an assumption of complete knowledge of the reasoner (e.g., \cite{BrennerN09}).

Solutions to the epistemic planning problem can be roughly divided into two classes: (1) the approach taken in \emph{Dynamic Epistemic Logic} (DEL) of using Kripke structures to maintain knowledge and then using models, such as event models, to update Kripke structures as events occur, for example, \citeasnoun{BolanderA11}; or (2) to maintain a set of formulae in a database and using belief update and revision to progress the database, for example, \citeasnoun{huang2017general}. In this paper, we adopt the latter approach.

We define epistemic planning for a particular class of problems in which the knowledge base is a \emph{proper epistemic knowledge base} (PEKB) \cite{Lakemeyer}. A PEKB is defined as finite set of epistemic literals, meaning that: (1) they do not allow for disjunctive belief; and (2) the depth of nested belief is bounded. We show that this provides certain theoretical guarantees on the computational complexity of progression of actions that have PEKBs as preconditions and effects, and directly offers an efficient means of planning using known methodologies.

We propose a means of encoding our definition of epistemic planning as a classical planning problem, enabling us to exploit state-of-the-art classical planning techniques to synthesize plans for these challenging planning problems.  A key aspect of our encoding is the use of \emph{ancillary conditional effects} -- additional conditional effects of actions which enforce desirable properties such as epistemic modal logic axioms (cf. Section \ref{sec:encoding}), and allow domain modellers to encode conditions under which agents are mutually aware of actions (cf. Section \ref{sec:ma-cond}).  By encoding modal logic axioms as effects of actions, we are using the planner to perform epistemic reasoning in addition to the standard reasoning about action and change. We implement this encoding as a compilation from epistemic planning problems to classical planning problems, and show that it handles a rich variety of problems.


Computational machinery for epistemic reasoning has historically appealed to theorem proving or model checking (e.g., \cite{van2004dynamic}), while epistemic planning, originally popularized within the DEL community, has shifted from a sole focus on theoretical concerns (e.g., \cite{LowePW11}) to a growing array of practical approaches and implementations. Previously, the problem of planning with epistemic goals has received limited attention in the automated planning community (e.g., \cite{Petrick:2006,bai-mom-mci-kr14} and most recently with multi-agent beliefs \cite{kominisbeliefs,huang2017general}). The work presented here is an important step towards leveraging state-of-the-art planning technology as a black-box sub-procedure to address rich epistemic planning problems of the sort examined by the DEL community. Indeed, we can readily solve existing examples in the DEL literature (cf. Section \ref{sec:evaluation}). We further discuss the relationship of our work to other modern epistemic planning work in Section \ref{sec:related}.

Parts of this paper have been published in earlier work, notably the theory of proper epistemic knowledge bases \cite{miller2016knowing,miller-ijcai16,muise-aamas-15} and parts of the encoding to classical planning \cite{muise2015planning}. However, the work in this paper extends that body of work in three key areas: (1) we provide a formal definition of epistemic planning over PEKBs, with earlier work \cite{muise2015planning} informally defining the problem; (2) we extend the encoding to deal with a restricted class of common knowledge, called \emph{always known}, which are propositions for which every agent will always know the value, such as static knowledge; and (3) we significantly expand the evaluation, defining new benchmark problems and presenting applications on which we have used our planner.

\label{page:model-separation}
After presenting the background notation required in Section \ref{sec:preliminaries}, we detail the syntax and semantics of the  Restricted Perspectival Multi-agent Epistemic Planning model in Section \ref{sec:rpmep}. We follow with the detailed encoding to classical planning in Section \ref{sec:encoding}. Extensions to greatly reduce the burden of modelling and improve planner efficiency are presented in Section \ref{sec:extensions} with grounded examples in Section \ref{sec:examples}. We investigate the empirical nature of our approach in Section \ref{sec:evaluation} and provide a discussion of related work and concluding remarks in Sections \ref{sec:related}-\ref{sec:conclusion}. For reader convenience, \ref{app:terms} provides a list of the common acronyms used throughout the paper. To illustrate the PDKB Description Language, a variant of PDDL \cite{pddlbook}, \ref{app:exemplary-domains} details the \emph{Grapevine} and \emph{Envelope} domains. In \ref{app:ancillary-domains} we demonstrate the encoding of ancillary conditional effects. Finally, for completeness, \ref{app:pekbs} outlines the theory of PEKBs in the context of related work, emphasising the nice logical and computational properties that make them a suitable representation for extending classical planning over belief bases, in a KD$_n$/KD45$_n$ context.

\begin{example}[Grapevine]
We will use a common example to explain the concepts introduced throughout the paper. Consider a scenario where a group of agents each have their own secret to (possibly) share with one another. Each agent can move freely between a pair of rooms, and broadcast any secret they currently believe to everyone in the room. Initially they only believe their own unique secret. Goals we might pose include the universal spread of information (everyone believes every secret), misconception (an agent holds a false belief about someone else's belief), etc. We will use $1,2,\cdots$ to represent the agents, and $s_1,s_2,\cdots$ to represent their secrets, respectively. Given our perspectival view of the setting, these may be cast either in terms of one of the acting agents in the environment, or as a third-party observer that has partial information about the world.
\end{example}

\section{Preliminaries}
\label{sec:preliminaries}

\subsection{Epistemic Logic}
\label{sec:del}

In this section, we introduce the concepts of epistemic logic, and in particular, the model of epistemic logic dealing with belief. We remain true to the terminology used in epistemic logic, so it is worth noting for those unfamiliar with the subject area that terms similar to those found in the planning literature are used. In particular, \emph{state} represents a configuration of what is true/false in the world, but \emph{reachability} is not defined in terms of actions applicability. Rather, it is a way to capture the space of possible states an agent believes to be possible.


Let $\P$ and $\agents$ respectively be finite sets of propositions and agents.
The set of well-formed formulae, $\L$, is obtained from the following grammar:
\[
	\phi ::= p \mid \phi \land \phi' \mid \belief{i} \phi \mid \neg \phi \mid \top \mid \bot
\]
in which $p \in \P$ and $i \in \agents$. $\belief{i} \phi$ should be interpreted as ``agent $i$ believes $\phi$'', and we will suppress the agent index when the formula holds for all agents. We will also use $\possible{} \phi$ as a syntactic shorthand for $\neg \belief{} \neg \phi$ and $\possible{i} \phi$ should be interpreted as ``agent $i$ considers it possible that $\phi$''.

The semantics is given using \emph{Kripke structures} \cite{Fagin}. Each Kripke structure is a tuple  $M = (\allworlds, \pi, R_1, \ldots R_n)$, in which $\allworlds$ is the set of all worlds considered in a model, $\pi \in \allworlds \rightarrow 2^{\P}$ is a function that maps each world to the set of propositions that hold in that world, and each $R_i \subseteq \allworlds \times \allworlds$ (for each $i \in \agents$) is a belief accessibility relation. %
%
Each relation $R_i$ captures the uncertainty of agent $i$ such that, given the actual world $w$, the set $R_i(w) = \set{w' \st R_i(w,w')}$ is the set of worlds that agent $i$ considers possible, i.e., indistinguishable from $w$.

Given these definitions, the satisfaction of a formula $\phi$ in a Kripke structure $M$ and a world $w$ is denoted as $M, w \vDash \phi$, and it is defined inductively over the structure of $\phi$:
\begin{center}
\begin{tabular}{llll}
 & $M, w \vDash \top$ & & \\
 & $M, w \nvDash \bot$ & & \\
 & $M, w \vDash p$ & iff &  $p \in \pi(w)$\\
 & $M, w \vDash \varphi \land \psi$ & iff & $M, w \vDash \varphi$ and $M, w \vDash \psi$\\
 & $M, w \vDash \neg \varphi$ & iff & $M, w \nvDash \varphi$\\
 & $M, w \vDash \belief{i}\varphi$ & iff & for all $v \in R_i(w)$,  $M, v \vDash \varphi$
\end{tabular}
\end{center}

We define entailment as: $\phi \vDash \psi$ if and only if for every model $M$ and world $w$ such that $M, w \vDash \phi$, we have $M, w \vDash \psi$. The pointed model $(M, w)$ defines an actual world in the model. 


As discussed by \citeasnoun{Fagin}, constraints on Kripke structures lead to particular properties of belief. If the Kripke structure is \emph{serial}, \emph{transitive}, and \emph{Euclidean} we obtain (arguably) the most common properties of belief:

\noindent
\begin{center}
\begin{tabular}{lll}
 $K$ & $\belief{}\phi \land \belief{}(\phi \supset \psi) \supset \belief{}\psi$ & (Distribution)\\
 $D$ & $\belief{} \phi \supset \possible{}\phi$ & (Consistency)\\
 $4$ & $\belief{}\phi \supset \belief{}\belief{}\phi$ & (Positive introspection)\\
 $5$ & $\possible{} \phi \supset \belief{} \possible{} \phi$ & (Negative introspection)
\end{tabular}
\end{center}

\noindent
These axioms collectively form the system referred to as \( \emph{KD45} \sub n \), where $n$ specifies that there are multiple agents in the environment. From the axioms, additional theorems can be derived. For example, in this work, we use the following theorems from \citeasnoun{hughes1996new} for reducing neighbouring belief modalities involving the same agent into a single belief modality:

\begin{center}
\begin{tabular}{rlrcrlr}
$\belief{i}\belief{i}\phi$     & $\equiv$ & $\belief{i}\phi$ & \hspace{3mm} & $\belief{i}\possible{i}\phi$ & $\equiv$ & $\possible{i}\phi$\\
$\possible{i}\belief{i}\phi$ & $\equiv$ & $\belief{i}\phi$ & & $\possible{i}\possible{i}\phi$    & $\equiv$ & $\possible{i}\phi$
\end{tabular}
\end{center}

\subsection{Proper epistemic knowledge bases}
\label{sec:background:pekbs}

Not surprisingly, reasoning (and planning) in these logical frameworks is computationally challenging \cite{Fagin,Aucher}. To mitigate this, \citeasnoun{Lakemeyer} define a \emph{proper epistemic knowledge base} (\kb) as a set of restricted formulae, called  \emph{restricted modal literals} (RMLs), of the form:
\[
	\phi ::= \top \mid \bot \mid p \mid \neg p \mid \belief{i} \phi \mid \possible{i} \phi 
\]
where \( p\in\P \) and $i \in \agents$. Thus, a \kb contains no disjunctive formulae.

The depth of an RML is defined as:
\begin{align*}
\emph{depth}(\phi) = \begin{cases}
1 +\emph{depth} (\psi) & \textit{if } \phi = \belief{i}(\psi) \\
1 +\emph{depth} (\psi) & \textit{if } \phi = \possible{i}(\psi) \\
0 & \textit{otherwise}
\end{cases}
\end{align*}
where \( p\in \P \). We will view a conjunction of RMLs equivalently as a set, and denote the set of all RMLs with bounded depth $d$ for a group of agents $\agents$ as $\allrmls(\P)$ (we drop the $\P$ qualifier when it is obvious from the context).

Because RMLs are represented as a sequence of $\belief{}$ and $\possible{}$ operators, ending in a propositional literal, they are in \emph{negation normal form} (NNF); i.e., negation appears only in front of propositional variables. Any standard modal literal with negations interleaved can be re-written into NNF using the equivalences,
\[ \neg \belief{i} \varphi \equiv \possible{i}\neg\varphi \mm{5} \neg\possible{i} \varphi \equiv \belief{i}\neg\varphi \mm{5} \neg\neg p \equiv p\]
%

To simplify exposition throughout the paper, we will take the negation of an RML to indicate its equivalent NNF (i.e., by using the repeated application of the above equivalences so that negations appear only at the literal level).

We use $Lit(\phi)$ to refer to the literal at the end of the RML $\phi$:
\[ Lit(\phi) = \begin{cases}
Lit(\psi) \mm{4} &\textrm{if } \phi = \belief{i}\psi \mm{2} \textrm{or} \mm{2} \phi = \possible{i}\psi \\
\phi &\textrm{otherwise}
\end{cases}
\]

\citeasnoun{Lakemeyer} show how to compile a \kb into \emph{prime implicate normal form} (PINF) in exponential time and space, and how to check entailment of this PINF formula in polynomial time. This compares to double-exponential time and space for non-restricted problems \cite{bienvenu09}. Thus, by sacrificing expressiveness, some computational cost can be reduced. Their entailment algorithm is sound for arbitrary formulae, and complete for PINF formulae.
The consistency of a PEKB is defined using the semantics of epistemic logic, where the PEKB is seen as the conjunction of the elements in the PEKB \cite{Lakemeyer}.

\citeasnoun{muise-aamas-15}  showed that for the logics \kd, compilation to PINF is not required: the \kb can be queried directly in polynomial time.
Querying is sound for any arbitrary \kd formula, and is complete for formulae in a particular normal form. 
\citeasnoun{miller-ijcai16} then further define belief update for PEKBs that can be calculated in polynomial time.


Extending to the KD45$_n$ case --- that is, adding positive and negative introspection (axioms 4 and 5 respectively) --- is straightforward using the equivalences in Section~\ref{sec:del}.
\citeasnoun{Lakemeyer} define an \emph{i-objective} formula as a formula that is about the world and agents other than $i$. For example, $\belief{j}(p \land \belief{i}\neg p)$ is $i$-objective, but $\belief{j}p \land \belief{i}\neg p$ is not. A formula is \emph{$i$-reduced} iff for all sub-formulae $\belief{i}\varphi$ and $\possible{i}\varphi$, $\varphi$ is $i$-objective.

One can see that any RML $\belief{i}\varphi$ or $\possible{i}\varphi$ can be $i$-reduced by repeatedly applying the equivalences above to strip out consecutive occurrences of modal operators of the same agent. Therefore, one can reduce both a KD45$_n$ PEKB and query into a \kd PEKB and query respectively, allowing application of the approaches in this section to KD45$_n$. We restrict our discussion in the remainder of this paper to the more simplified case of \kd, assuming that for \kdff, consecutive modal operators of the same agent have been removed.

\paragraph{Expressivenes of PEKBs} Given that PEKBs restrict more general epistemic logic by removing disjunction, it raises the question whether they are too restrictive. The answer depends on which perspective is taken. On the one hand, omitting disjunction from the language means that some interesting properties around disjunctive preconditions and goals need to be encoded manually, and this manual encoding needs to carefully ensure all logical properties are maintained. On the other hand, contemporary classical and non-deterministic planning tools typically do not support disjunction on even propositional formulae, so expanding the planning language to a PEKBs does not limit the planning at all -- it makes modelling some problems more straightforward. In Section~\ref{sec:eval-hattari}, we highlight an example of where the lack of handling disjunctive knowledge limits what we can represent. Further, we dive deeper on the expressive differences between our approach and related works in Section~\ref{sec:related}.

PEKBs have been shown to be expressive enough for many applications, such as collaborative filtering \cite{Lakemeyer} and team formation \cite{muise2015towards}. The key contribution of this paper is to show how epistemic planning can be done \emph{efficiently}. What is clear is that to solve planning problems efficiently, trade-offs in expressiveness must be made. The inclusion of disjunction would immediately rule out the approach of compiling to classical or non-deterministic planning without explicitly reifying sub-formulae. Importantly, if classical planners did support disjunction, then including disjunctive epistemic knowledge bases would require a double-exponential compilation step just on the epistemic formulae themselves \cite{bienvenu08}, rather than polynomial.

As the results in this paper later show, we can indeed solve several standard benchmarks in epistemic planning more efficiently than other epistemic planners (both for expressiveness and computational reasons). In our view, to advance the field of epistemic planning to the point where planners are viable for solving large-scale planning tasks rather than epistemic puzzles, restricted representations like PEKBs will be essential.


\subsection{Classical and FOND Planning}
\label{sec:classical-fond-planning}
A classical planning problem consists of a tuple $\la \fluents, I, G, \operators \ra$, where $\fluents$ is a set of fluent atoms, $I$ is the initial state, $G$ is the goal, and $\operators$ is a set of operators. A \emph{complete state} (or just \emph{state}) $s$ is a subset of the fluents $\fluents$ with the interpretation that fluents not in $s$ are false (equivalently, this can be seen as a conjunction between the literals found in $s$ and the negated literals not found in $s$). A \emph{partial state} $s$ is similarly a subset of the fluents $\fluents$, but has the interpretation that literals not found in $s$ can take on any value (thus equivalent to a conjunction of just the literals in $s$). $I$ is a complete state while $G$ is a partial state. Every operator $o \in \operators$ is a tuple $\la \pre_o, \poseffs{o}, \negeffs{o} \ra$, and we say that $o$ is applicable in $s$ iff $\pre_o \subseteq s$. 
The set $\poseffs{o}$ (resp. $\negeffs{o}$) contains conditional effects describing the fluent atoms that should be added (resp. removed) from the state when applying the operator. Finally, every conditional effect in $\poseffs{o}$ or $\negeffs{o}$ is of the form  $\condeff{\cond}{\lit}$ where $\cond$ is the condition for the effect and $\lit$ is a fluent that is the result of the effect. The condition $\cond$ consists of a tuple $\la \condp, \condn \ra$ where $\condp$ is the set of fluents that must hold and $\condn$ the set of fluents that must not hold. A conditional effect $\condeff{\la \condp, \condn \ra}{\lit}$ \emph{fires} in state $s$ iff $\condp \subseteq s$ and $\condn \cap s = \emptyset$. Assuming $o$ is applicable in $s$, and $\poseffs{o}(s)$ (resp. $\negeffs{o}(s)$) are the positive (resp. negative) conditional effects that fire in state $s$, the state of the world $s'$ after applying $o$ is defined as follows:
\begin{align*}
s' = s &\setminus \set{\lit \st \condeff{\cond}{\lit} \in \negeffs{o}(s)} \\
&\cup \set{\lit \st \condeff{\cond}{\lit} \in \poseffs{o}(s)}
\end{align*}

Our account of classical planning mirrors the standard representation (see, for example, \cite{ghallab2004automated}), with the exception that we make explicit the fluent atoms that are added, deleted, required to be in, or required to be absent from the state of the world. This simplifies the exposition when we encode  nested beliefs as a classical planning problem.

Fully-observable non-deterministic (FOND) planning extends classical planning with non-deterministic operators. That is, operators can have one \emph{or more} effects, and exactly one of these effects occurs when the action is executed. The planning agent does not know which effect will occur prior to execution, but can observe which effect occurred immediately after execution. To extend classical planning, instead of each operator containing a single pair of positive and negative effects, an operator has a \emph{set} of non-deterministic effects, with each containing positive and negative effects: $\langle Pre_o, \Eff_o\rangle$, in which every non-deterministic effect $E \in \Eff_o$ consists of two sets of conditional effects, $\poseffs{E}$ and $\negeffs{E}$. A solution to a FOND planning problem is a \emph{policy} $\pi$, which maps states to operators. A policy represents a set of possible sequences of actions. If every sequence in a policy reaches the goal, it is said to be \emph{strong}; otherwise, if only some sequences reach the goal, it is \emph{weak}.

\section{Restricted Perspectival Multi-agent Epistemic Planning}
\label{sec:rpmep}

In this section, we define the syntax and semantics of \emph{restricted perspectival multi-agent epistemic planning}  (\epplan). At a high-level, this is defined simply as planning over states of the world where states are PEKBs, instead of collections of propositional fluents. The choice of PEKBs is suitable for two reasons:

\begin{enumerate}
    \item The syntactic restrictions imposed by PEKBs that prevent disjunction and infinite nesting are consistent with the syntactic restrictions employed by STRIPS-based planners, in which arbitrary disjunction is not permitted, and the set of literals (fluents) in a planning problem is finite. In this sense, using PEKBs increases the expressiveness of the STRIPS language to include epistemic formulae.
    \item PEKBs come with nice logical and computational properties, as we outline in this section, and elaborate in \ref{app:pekbs}. First, a \emph{consistent} PEKB --- that is, one with no contradictory statements --- is \emph{logically separable}, which means the literals in the knowledge base do not interact to produce new formulae. Further, a consistent PEKB can be queried in polynomial time without a pre-compilation step such as the one used by \citeasnoun{Lakemeyer}, and can be queried in constant time with an exponential pre-compilation step. Finally, a consistent PEKB can be updated with new literals and remain consistent using a polynomial-time algorithm.
\end{enumerate}

We assume that the state represents the mental model of a particular agent that perceives an environment that includes all other agents. All reasoning is from the perspective of this single agent. The fluents that are true in a state correspond to the RMLs that the agent believes, while the fluents that are false correspond to the RMLs that the agent does not believe.
Action execution, then, is predicated on the agent \emph{believing} that the preconditions are satisfied. Similarly, the mental model of the agent is updated according to the effects of an action.
Note that we do not need to enforce a separation of ontic and epistemic effects -- the same action can update belief about propositions as well as RMLs. This is due to the interpretation that the state of the world represents the mental model of a given agent: every effect is epistemic in this sense.

\subsection{Syntax}
\label{sec:syntax}

In describing the actions for a \epplan problem, we consider both conditional and non-deterministic effects; combined they allow for a rich variety of epistemic problems to be described. The non-determinism, however, is of a restricted form in that it is fully observable, meaning that the executing agent does not know what the outcome of an action will be, however, it will observe the outcome immediately after executing the action. While we hope to remove this restriction in the future, it allows us to account for various contingencies during the planning process, and model aspects such as questions asked to other agents in the environment.

The basic atomic action in RP-MEP is a PEKB planning action. Essentially, an action is just a classical planning action, but in which the preconditions and effects can be PEKBs instead of just propositions.


\begin{definition}{\epplan Planning Action}
	\label{defn:rp-mep-planning-action}
A \epplan Planning action $a$ is a pair  \(\la \pre_a, \Eff_a \ra\), in which $\pre_a$ is a PEKB capturing the preconditions of $a$ and $\Eff_a$ is the set of non-deterministic effects of $a$, defined as follows. Every non-deterministic effect $E \in \Eff_a$ is constituted by a set of \emph{conditional effects} \(  \{(\gamma\sub 1, \literala\sub 1), \ldots, (\gamma\sub k, \literala\sub k)\} \), in which each \( \gamma\sub i \) is a PEKB called the \emph{condition} of the conditional effect, and each \( \literala\sub i \) is a RML called the \emph{effect} of the conditional effect. A conditional effect \( (\gamma\sub i, \literala\sub i)\) is  read informally to say that if the condition is true before the action executes, then the effect will be true after the action executes. 
\end{definition}

\noindent
We can now define a planning problem as follows:

\begin{definition}{Multi-Agent Epistemic Planning Problem}
A \emph{\textbf{m}ulti-agent \textbf{e}pistemic \textbf{p}lanning} (\fullepplan) problem
\label{def:ma-ep-plan}
\( \D \)  is a tuple of the form \(  \la \P, \A, \agents, \init, \goal \ra \), where \( \P \) and \( \agents \) are as above, \( \A \) is a set of \epplan planning actions, $\init$ is the \kb representing the \emph{initial theory} and $\goal$ is a \kb capturing the \emph{goal condition.}
\end{definition}





Following \citeasnoun{Reiter} and \citeasnoun{Ditmarsch}, the above action formalization can be expressed as standard \emph{precondition} and \emph{successor state axioms}, which would then define the meaning of achieving the goal. Using PDL-like syntax, we interpret $[\alpha]\goal$ as ``$\goal$ holds after executing action $\alpha$''. We formally define this below in Section \ref{sec:plan-ver-and-gen}.

We define a \emph{\textbf{r}estricted \textbf{p}erspectival} multi-agent epistemic planning problem (\epplan problem) for depth bound $d$ and the root agent $\star \in \agents$ as a \fullepplan problem with the additional restrictions that: (1) every RML is from the perspective of the root agent -- i.e., it is from the following set:
\[
\set{\acting \literala \st \literala \in \allrmls} \cup \set{\possible{\actingsym} \literala \st \literala \in \allrmls},
\]
\noindent and (2) there is no disjunctive belief: the initial theory, goal specification, and every precondition are all PEKBs, every effect is a single RML, and every effect condition is a PEKB.

We address the planning problem from the view of an acting agent, where the designated root agent $\star$ is the one for which we plan. Intuitively, this means that conditional effects are formulated in the context of the root agent; e.g., we would have a conditional effect of the form $(\set{\acting\gamma}, \acting \literala)$ for action $a$ in a \epplan problem to capture the fact that the root agent will believe $\literala$ if it believed $\gamma$ before $a$ occurred (as formalised in the next section).

This admits a rich class of planning problems; e.g., it is reasonable to assume that the root agent's view of the world differs from what a particular agent $i$ believes, and so another conditional effect of $a$ might be $(\set{\acting \gamma}, \acting \belief{i} \neg \literala)$ -- even though the root agent believes doing $a$ would make $\literala$ true if $\gamma$ holds, the root agent believes that $i$ will believe $\neg \phi$ if $\gamma$ holds. In particular, this is easily shown to generalize a standard assumption in the non-epistemic multi-agent planning literature \cite{Wen} that all agents hold the same view of what changes after actions occur.

In the next section, we show how a restricted perspectival multi-agent epistemic planning problem can be represented as a classical planning problem, where the key insight is to encode reasoning features (such as deduction in \kd) as \emph{ramifications} realized using ordinary planning operators.

\subsection{Semantics}

To formally define the semantics of epistemic planning over \kb{s}, we use the notation defined by \citeasnoun{miller-ijcai16} for belief update in PEKBs. The theoretical justification for this process is explored in \ref{app:pekbs}, but for clarity we restrict the rhetoric in this section to only those concepts required to describe our encoding.

\subsubsection{Action Progression with PEKBs}

While syntactic treatments of belief and knowledge have been proposed in the past; e.g.\ \cite{eberle1974logic,konolige1983deductive}, PEKBs place restrictions on the syntax of formulae to provide desirable computational properties. The key result by
\citeasnoun{Lakemeyer} is that by eliminating disjunction from formulae, PEKBs can be compiled in into a set of prime implicates, similar to the method employed by \cite{bienvenu08,bienvenu09} for the logic $K_n$. This allows entailment queries to be answered in polynomial time by structurally traversing the prime implicates instead of querying the original belief base. The cost is that the compilation into prime implicates is exponential, rather than double exponential in the case of epistemic logic with disjunction. One could reason efficiently (polynomial time) at the level of prime implicates, but we opt to further process the PEKBs so that they are deductively closed -- this will greatly simplify the propositional encoding later in Section \ref{sec:encoding}.

In this section, we define the semantics of epistemic planning over PEKBs. For this reason, in order to keep the terminology intuitive, we call these \emph{PEKB states}. For simplicity, we define a PEKB state $P$ as a \emph{deductively-closed} subset of formulae. We use $\negkb{P}$ to denote the \kb that contains the negation of every RML in $P$; that is
$\negkb{P} = \set{\neg \literala \st \literala \in P}$.
%
Following Definitions \ref{defn:rp-mep-planning-action} and \ref{def:ma-ep-plan}, we can define the procedures to modify \kbs as follows:

\begin{definition}{Belief update and erasure in \kbs}
\label{defn:belief-update-and-erasure}
Given deductively-closed \kbs $P$ and $Q$, we define $P \erase Q$ and $P \update Q$  as the \emph{belief erasure} and \emph{belief update} of  $P$ and $Q$ respectively, as follows:
\begin{align*}
P \erase Q &=  P \setminus Q\\
P \update Q &= (P \erase \negkb{Q}) \cup Q.
\end{align*}
Belief update of $P$ with $Q$ is defined as the standard `forget (erase) then conjoin': anything in $P$ disagreeing with $Q$ is forgotten, then everything from $Q$ is added.
\end{definition}

\citeasnoun{miller-ijcai16} present definitions of belief update and erasure for PEKBs and evaluate them against the \citeasnoun{KM91} postulates for belief update \cite{KM91}. They also show that belief update and erasure in PEKBs can be computed in polynomial time. 

\begin{definition}{\epplan Progression}
We define the progression of a single deterministic effect $E$ applied to a deductively-closed PEKB state $P$, labelled $\progr{P}{E}$, as:
\begin{align*}
\progr{P}{E}\ =\ & (P \erase (R \cup U)) \update Q\\
 Q\ =\ & \bigcup_{(\gamma, \literala) \in E} \{\literalb \mid \gamma \subseteq P \land \literala \vDash \literalb \}\\
 R\ =\ & \bigcup_{(\gamma, \neg \literala) \in E} \{\literalb \mid \gamma \subseteq P \land \literala \vDash \literalb \}\\
 U\ =\ & \bigcup_{(\gamma, \literala) \in E} \{\neg \literalb \mid \negkb{\gamma} \cap P = \emptyset \land \neg\literala \vDash \neg\literalb \}.
\end{align*}
\noindent $Q$ defines the set of literals to be added, $R$ defines the set of literals to be deleted, while $U$ defines the set of \emph{uncertain firing} literals to be deleted. Uncertain firing captures instances in which an agent is unsure whether a conditional effect is true, denoted $\negkb{\gamma} \cap P = \emptyset$. If an agent is unsure, then it should not believe the effect (unless it was already true), but must admit that it could be true. Therefore, it must not believe the opposite, and we should remove $\neg \literala$ and anything that follows from it deductively.
\label{defn:rpmep-progression}
\end{definition}



From this, we can define the set of possible progressions of a non-deterministic \epplan action $a=\la \pre_a, \Eff_a \ra$ in a deductively-closed PEKB state $P$:
\[
\progr{P}{a} \myisdef \set{\progr{P}{E} \st E \in \Eff_a \mm{1} \mathrm{and} \mm{1} \pre_a \subseteq P}
\]

Intuitively, $\progr{P}{a}$ is defined as a set of PEKB states, each corresponding to the application of a non-deterministic effect of the \epplan action $a$. 
In case the \epplan action $a$ is not executable in $P$, i.e. $\pre_a \not\subseteq P$, then the set is empty. 

A (\epplan) \emph{policy} is a function $\alpha : \P \rightarrow \A$ mapping PEKB states to \epplan actions ($\P$ is used to denote the set of all possible PEKB states). Valid policies can be partial functions, because their terminating states can be undefined.
%

A \emph{trajectory} is a sequence of PEKB states, and we say that a trajectory $P_0,\ldots, P_n$ is  \emph{induced} by a policy $\alpha$ 
from a PEKB state $P$ iff $P=P_0$ and for $i\in[0,n-1]$ we have that $\alpha(P_i)$ is defined 
and that $P_{i+1}\in\progr{P_i}{\alpha(P_i)}$. 
Intuitively, as actions have non-deterministic effects, there are in general many trajectories which are induced by the same policy. 

A PEKB state $P$ is said to be an \emph{end state} of a policy $\alpha$ from a PEKB state $P_0$, denoted $P\in end(P_0,\alpha)$, iff there exists a trajectory $P_0,\ldots,P$ induced by $\alpha$ from $P_0$ such that $\alpha(P)$ is either not defined or empty.

We further assume that a valid policy does not yield inescapable cycles. In other words, for every trajectory $T$ induced by policy $\alpha$ from a PEKB $P$, there exists a trajectory $T'$ so that $T\cdot T'$ is induced by $\alpha$ from $P$ and the last state of $T\cdot T'$ is an end state of $\alpha$. This yields policies that correspond to strong cyclic plans from FOND planning.
\label{page:cyclic-policies}

It is worth pointing out the implicit assumption of ``fairness''. Loosely speaking, fairness refers to a property of the non-determinism in an environment that captures the notion of all possible outcomes having some chance of success: if the agent finds themselves repeatedly in the same situation and executes the same action, every possible outcome should occur infinitely often in the limit. Due to our definition of a valid policy, we are assuming fairness. This may be an abstraction of the true nature of the environment (e.g., a non-deterministic action that ``asks'' another agent a question is inherently unfair), but it is still a useful abstraction for reasoning with both uncertainty in action outcomes and uncertainty in agent beliefs. There are encoding techniques to solve unfair non-deterministic planning problems with solvers assuming fairness (e.g., \cite{CamachoM16}), but those notions are out of scope for this work.


\subsubsection{Plan Verification and Generation}
\label{sec:plan-ver-and-gen}

In this section, we formally define the problems of plan validation and plan verification for \epplan.

Given a conjunction of RMLs $\phi$, we use $[\alpha]\phi$ to represent that $\phi$ holds in all trajectories defined by policy $\alpha$, given some start state. Formally, this is defined as:
\begin{center}
\begin{tabular}{llll}
 & $P \vDash [\alpha]\phi$ & iff & $Q \vDash \phi$ for all $Q \in end(P,\alpha)$,
\end{tabular}
\end{center}
in which $P$ is a deductively-closed PEKB. If $[\alpha]\phi$ holds in $P$, then $\phi$ holds for all end states of the policy $\alpha$ from $P$ and $\alpha$ is said to be a \emph{strong} policy for $\phi$ from $P$.  
We define $\langle\alpha\rangle \phi$ as shorthand for $\neg [\alpha]\neg \phi$, which means that $\phi$ holds on at least one end trajectory of $\alpha$. If $\langle\alpha\rangle \phi$, then $\alpha$ is said to be a \emph{weak} policy for $\phi$ from $P$. 

%
The notions of plan verification and generation rely on the idea of a \emph{valid policy} for a goal: a policy that achieves the goal from the initial state. Formally, we have the following type of valid policies.

\begin{definition}{Plan Verification and Generation}
Given an \epplan problem, $\la \P, \A, \agents, \init, \goal \ra$, a policy $\alpha$ is a valid \emph{weak} policy for $\goal$ iff the following holds:
%
\[ \init \models \langle\alpha\rangle\goal \]
A policy $\alpha$ is a valid strong policy for $\goal$ iff the following holds:

\[ \init \models [\alpha]\goal \]

%
%
%

The task of \emph{plan verification} is to determine if a policy is a valid (weak or strong) policy for $\goal$. The task of \emph{plan generation} is to generate such a policy.
\end{definition}

Our notion of plan generation and verification follows closely that of traditional fully-observable non-deterministic planning, only with our states of the world being represented by PEKBs in lieu of an assignment of true/false to the fluents in the domain. It is unique in relation to other formalisms of epistemic planning, in that our restriction to PEKBs is not an assumption that the other related systems adopt. The more fundamental differences are discussed later in Sections \ref{sec:evaluation} and \ref{sec:related}.

\section{Propositional Encoding of \epplan}
\label{sec:encoding}

In this section, we show how to encode an \epplan into a standard propositional planning problem, as either classical planning or FOND planning as required.
The novelty in this work is that, given an \epplan planning problem, we convert this into a propositional planning problem that can be solved with any off-the-shelf classical/FOND planning tool that supports conditional effects. As FOND planning is a generalisation of classical planning, we define our encoding for FOND planning only.

The basic framework that we take is as follows. For every RML in the domain problem, create a standard propositional fluent to represent that RML. Then, for every action in the original problem, replace each RML with its propositional fluent, and add new conditional effects to the action to handle the semantics of \epplan. The result is a FOND planning problem in which a solution to the FOND problem is a solution for the \epplan problem.

The remainder of the section will proceed as follows:
\begin{enumerate}
    \item We present the framework for our encoding of an \epplan problem into FOND planning.
    \item We specify how the belief update operator defined in Section~\ref{sec:rpmep} is encoded.
    \item We specify how to  correctly update if an agent is uncertain whether a conditional effect fires.
\end{enumerate}

\subsection{Base Encoding}

 First, using the theorems for reducing neighbouring modalities outlined in Section~\ref{sec:del}  (e.g.\ $\belief{i}\belief{i}\phi \equiv \belief{i}\phi$), remove any neighbouring modalities from each RML that appears in any part of the planning problem, including preconditions, effects, fluents, etc.\ This preserves the semantics of the original problem, but simplifies the encoding.

Then, we transform the problem into an equivalent propositional planning problem.
Intuitively, every RML in $\allrmls$ will correspond to a single fluent in $\fluents$ (e.g., both $\belief{1} p$ and $\possible{1} \neg p$ will become fluents), and the operators will describe how the mental model of our root agent should be updated. Formally, we define the encoding of a \epplan problem as follows:





%
%
\begin{definition}{Encoding of \epplan}
\label{def:encoding}
Let $\invertedbelief{i}$ and $\ninvertedbelief{i}$ be functions that map $i$'s positive (resp. negative) belief from a PEKB $P$ to the respective fluents:
\begin{align*}
\invertedbelief{i}(P) = \set{\lit_\phi \st \belief{i} \phi \in P} \\
\ninvertedbelief{i} (P) = \set{\neg\lit_\phi \st \neg \belief{i} \phi \in P}
\end{align*}

Given a \epplan problem, \(  \la \P, \A, \agents, \init, \goal \ra \) and a bound $d$ on the depth of nested belief we wish to consider, we define the propositional encoding as the tuple $\la \fluents, I, G, \operators \ra$ such that:
\begin{align*}
F \myisdef \set{l_\phi \st \phi \in \allrmls} \mm{8} I \myisdef \invertedbelief{\actingsym}(\closure{\init}) \mm{8} G \myisdef \invertedbelief{\actingsym}(\goal)
\end{align*}
\noindent and for every action $\la \pre_a, \Eff_a \ra$ in $\A$, we have a corresponding operator $\la \pre_o, \Eff_o \ra$ in $\operators$ such that:
\begin{align*}
\pre_o &\myisdef \invertedbelief{\actingsym}(\pre_a) \\
\Eff_o &\myisdef \set{\la \poseffs{E}, \negeffs{E} \ra \st E \in \Eff_a} \\
\poseffs{E} &\myisdef \set{\condeff{\la \invertedbelief{\actingsym}(\gamma), \overline{\ninvertedbelief{\actingsym}(\gamma)}\ra}{\lit_\phi} \st (\gamma, \acting \phi) \in \mathit{E}} \\
\negeffs{E} &\myisdef \set{\condeff{\la \invertedbelief{\actingsym}(\gamma), \overline{\ninvertedbelief{\actingsym}(\gamma)}\ra}{\lit_\phi} \st (\gamma, \neg \acting \phi) \in \mathit{E}}
\end{align*}
Whenever clear from the context, we will use $\poseffs{o}$ and $\negeffs{o}$ to refer to the effects associated with an action with a single outcome.
\end{definition}

\subsection{Maintaining Logical Closure}

Because of the direct correspondence, we will use the RML notation and terminology for the fluent atoms in $\fluents$. The encoding, thus far, is a straight-forward adaptation of the \epplan definition that hinges on two properties: (1) there is a finite bound on the depth of nested belief; and (2) we restrict ourselves to representing RMLs and not arbitrary formulae. Crucially, however, we wish to maintain the assumption that the agents are internally consistent with respect to \kd, and would like to do so in a pragmatic/syntactic manner. To accomplish this, we define a systematic closure procedure, $\closure$, that deduces a new set of RMLs from an existing one under \kd, allowing us to pre-compute all of the necessary inferences \emph{in advance of planning}:

\begin{definition}{RML Closure}
\label{def:closure}
Given an RML $\lit$, we define $\closure(\lit)$ to be the set RMLs that are \kd logical consequences of $\lit$. $\closure(\lit)$ is computed by repeatedly applying the $D$ axiom ($\belief{i}\psi \supset \possible{i}\psi$) to the RML, resulting in the set of all RMLs that follow logically from $\phi$ using the $D$ axiom. This can be done by simply replacing all combinations of occurrences of $\belief{i}$ with $\possible{i}$; e.g., for the RML $\belief{i}\possible{j}\belief{k}p$, the resulting set would be $\{\possible{i}\possible{j}\belief{k}p,~ \belief{i}\possible{j}\possible{k}p,~ \possible{i}\possible{j}\possible{k}p\}$.

Similarly, the closure of a PEKB is simply the union of the closure of its elements, i.e., $\closure(P) = \bigcup_{\lit \in P}\closure(\lit)$.
\end{definition}



  


Given the finite nature of a single RML, the closure is also finite.

\begin{theorem}[Soundness \& completeness of $\closure{}$]\label{thm:correct-closure}
Given a PEKB $P$ in KD$_n$, $\closure(P) = \{\literala \mid P \vDash \literala \textrm{, where } \literala \textrm{ is an RML}\}$.
\end{theorem}

\begin{proof}
As shown by \citeasnoun{miller-ijcai16}, a PEKB in KD$_n$ is \emph{logically separable}. Intuitively, this means that we cannot infer any new RML from a PEKB by combining two RMLs taken from the PEKB that we cannot already infer from just one RML. Therefore, to compute $\closure(P)$ we can just apply the axioms K and D to individual RMLs in the PEKB $P$. The repeated application of axioms K and D to each of the RMLs in the PEKB, when computing $\closure{}$, is a sound inference. Note that only a finite number of applications are possible given the finite nesting of each RML and the finite size of the PEKB.

For completeness, note that because any PEKB $P$ is logically separable, we can compute $\{\literala \mid P \vDash \literala\}$, where $\literala$ is an RML, by taking the union of $\{\literala \mid \lit \vDash \literala\}$ for each $\lit \in P$. We can then prove via induction that each $\literala$ will be in $\closure(\lit)$:

Case $\lit \equiv \belief{i}\psi$: Axiom D results in $\possible{i}\psi$, which is an RML because $\psi$ is an RML. Further, $\possible{i}\psi$ is in $\closure(\lit)$ from the definition of $\closure{}$ (Definition~\ref{def:closure}). By induction,  any further application of \kd axioms to $\possible{i}\psi'$ are in $\closure(\lit)$. Because $\psi$ is itself an RML and cannot contain disjunction or implication, axiom K does not apply. 

Case $\lit \equiv \possible{i}\psi$: Axiom D does not apply directly, and by induction, further application of \kd axioms to $\possible{i}\psi$ will be in $\closure(\lit)$. As with the previous case, axiom $K$ does not apply.

Therefore, $\closure{}$ is both sound and complete.
\end{proof}


Along with the requirement that an agent should never believe an RML and its negation, we have two further constraints on the PEKBs that we are encoding:
\begin{align*}
\phi \in s &\Rightarrow \neg \phi \notin s \\
\phi \in s &\Rightarrow \forall \psi \in \closure(\phi), \psi \in s
\end{align*}

The enforcement of such state constraints can either be achieved procedurally within the planner, or representationally.  We choose the latter, appealing to a solution to the well-known ramification problem (e.g., \cite{Pinto99,LinR94}), representing these state constraints as \emph{ancillary conditional effects} of actions that enforce the state constraints.  The correctness of the resulting encoding is predicated on the assumption that the domain modeller provided a consistent problem formulation (i.e., there are no inherent inconsistencies in the goal, initial state, or action effects). This mirrors the common assumption in modelling for planning more generally.
The ancillary conditional effects for operator $o$ are as follows:
%
\begin{align*}
\condeff{\cond}{\lit} \in \poseffs{o} &\Rightarrow \condeff{\cond}{\neg \lit} \in \negeffs{o} 
\stepcounter{equation}\tag{\theequation}\label{eq:remove-negation} \\
\condeff{\cond}{\lit} \in \poseffs{o} &\Rightarrow \forall \lit' \in \closure(\lit),\hspace{2mm}
\condeff{\cond}{\lit'} \in \poseffs{o}
\stepcounter{equation}\tag{\theequation}\label{eq:closure}
\end{align*}

\begin{example}
Returning to our example, consider the effect of agent 1 telling secret $s_1$ to agent 2. Assuming there is no positive or negative condition for this effect to fire, the effect would be $\condeff{\la \emptyset, \emptyset \ra}{\belief{2} s_1} \in \poseffs{}$. Using~(\ref{eq:remove-negation}) would create $\condeff{\la \emptyset, \emptyset \ra}{\neg \belief{2} s_1} \in \negeffs{}$ and (\ref{eq:closure}) would create $\condeff{\la \emptyset, \emptyset \ra}{\neg \belief{2} \neg s_1} \in \poseffs{}$. Subsequently, (\ref{eq:remove-negation}) would fire again creating $\condeff{\la \emptyset, \emptyset \ra}{\belief{2} \neg s_1} \in \negeffs{}$. We can see already, with this simple example, that effects may cascade to create new ones.
\end{example}

The second issue is to ensure the state remains consistent under \kd when removing beliefs. If we remove an RML $\lit$, we should also remove any RML that could be used to deduce $\lit$. To compute the set of such RMLs, we use the contrapositive: $\neg \lit'$ will deduce $\lit$ if and only if $\neg \lit$ deduces $\lit'$ (i.e., $\lit' \in \closure(\neg \lit)$). We thus have the following additional conditional effects for operator $o$:
\begin{align}
\condeff{\cond}{\lit} \in \negeffs{o} \Rightarrow \forall \lit' \in \closure(\neg \lit),\hspace{1mm} \condeff{\cond}{\neg \lit'} \in \negeffs{o}
\label{eq:uncertainty-closure}
\end{align}

\begin{example}
Consider the effect of an action informing us that agent 2 should no longer believe that agent 1 does not believe agent 2's secret: $\condeff{\la \emptyset, \emptyset \ra}{\neg \belief{1} s_2} \in \negeffs{}$. Using (\ref{eq:uncertainty-closure}), we would have the additional effect $\condeff{\la \emptyset, \emptyset \ra}{\belief{1} \neg s_2} \in \negeffs{}$. If $\belief{1} \neg s_2$ remained in our knowledge base, then so should $\neg \belief{1} s_2$ assuming that our knowledge base is deductively closed.
\end{example}

The effect of these two rules together is to encode the problem such that entailment is effectively done by computing the deductive closure of the PEKB each time an action effect is evaluated, matching the definition of progression in Section~\ref{sec:rpmep}. All possible RMLs are pre-compiled into a planning problem, and during planning, these are used multiple times, meaning that the pre-compilation is a valuable step. Given the indexing used in most modern planners, entailment for checking preconditions and effect conditions in planning actions becomes a constant-time operation.


\subsection{Uncertain Firing}

To complete the faithful transformation of a \epplan problem to a FOND problem, we must also consider the third case of \epplan progression defined in Definition~\ref{defn:rpmep-progression}: when an agent is uncertain whether a conditional effect should fire due to the uncertainty of beliefs. 

To encode this, we appeal to a common technique in planning under uncertainty (e.g., \cite{PetrickL02,PalaciosG09}): when the conditions of a positive conditional effect are not believed to be false, the negation of the effect's result can no longer be believed. Intuitively, if an agent is unsure whether a conditional effect fires then it must consider the condition's effect possible, and thus no longer believe the negation of the effect. We create the following additional conditional effects:
\begin{align}
\condeff{\la \condp, \condn \ra}{\lit} \in \poseffs{o} &\Rightarrow \notag\\ \condeff{\la \emptyset, \set{\neg \phi \st \phi \in \condp} &\cup \condn \ra}{\neg \lit} \in \negeffs{o}
\label{eq:uncertainty-ce}
\end{align}

\begin{example}
\label{ex:trustworthy}
Consider a conditional effect for the action of agent 1 sharing their secret that stipulates that if agent 2 believes that agent 1 is trustworthy (denoted as $\belief{2}t_1$), then agent 2 would believe agent 1's secret: $\condeff{\la \set{\belief{2}t_1}, \emptyset \ra}{\belief{2}s_1} \in \poseffs{}$. Using (\ref{eq:uncertainty-ce}), we would derive the new negative effect $\condeff{\la \emptyset, \set{\neg \belief{2}t_1} \ra}{\neg \belief{2}s_1} \in \negeffs{}$. Rule~\ref{eq:uncertainty-closure} would then fire, which would remove $\belief{2}\neg s_1$. Intuitively, although agent 2 is unsure about agent 1's trustworthiness, it should no longer believe that the secret is false.
\end{example}







In what follows, we denote by $s'=result(s,\la \poseffs{o},\negeffs{o} \ra)$ the new state resulting from the application of a non-deterministic effect $\la \poseffs{o},\negeffs{o} \ra$ of an operator $o$, as formally defined in Section~\ref{sec:classical-fond-planning}. With this notation at hand, we give the following main result. 

\begin{theorem}
\label{thm:belief-update}
Given an \epplan action $a=\la \pre_a, \Eff_a \ra$ and $E\in\Eff_a$, let $\la \pre_o, \Eff_o \ra$ be the operator corresponding to $a$ and $E_o=\la \poseffs{E},\negeffs{E} \ra$ its conditional effects corresponding to $E$, which are obtained according to Definition~\ref{def:encoding} and rules (\ref{eq:remove-negation})-(\ref{eq:uncertainty-ce}). Then, 
for any consistent PEKB $P$ we have that 
$Q=\progr{P}{E}$ iff $\closure{(\invertedbelief{\actingsym}(Q)\cup \ninvertedbelief{\actingsym}(Q))}=result(\invertedbelief{\actingsym}(P)\cup \ninvertedbelief{\actingsym}(P),E_o)$.  
\end{theorem}

\begin{proof}
First, recall that $I \myisdef \invertedbelief{\actingsym}(\closure{(\init)})$ by Def.~\ref{def:encoding}, hence $\closure{(\invertedbelief{\actingsym}(\init)\cup \ninvertedbelief{\actingsym}(\init))} =  I$ as $\ninvertedbelief{\actingsym}(\init)$ is empty for any \epplan problem (the initial theory is restricted to positive RMLs only -- see Section~\ref{sec:syntax}). Also, note that the encoded FOND problem is such that states are in fact always closed, hence we can consider $\closure(\invertedbelief{\actingsym}(P)\cup \ninvertedbelief{\actingsym}(P))$ in the above theorem. Consider a $l_\phi$ for which the equality in the theorem does not hold. Since by Def.~\ref{def:encoding} $l_\phi\in \invertedbelief{\actingsym}(P)$ iff $\acting \phi\in P$ (and similarly for $\ninvertedbelief{\actingsym}$), Theorem~\ref{thm:correct-closure} implies there are four cases to consider, which we prove by contradiction.

Assume $l_\phi$ is both in $\closure(\invertedbelief{\actingsym}(P)\cup \ninvertedbelief{\actingsym}(P))$ and in  $result(\invertedbelief{\actingsym}(P)\cup \ninvertedbelief{\actingsym}(P),E_o)$, but not in  $\closure(\invertedbelief{\actingsym}(Q)\cup \ninvertedbelief{\actingsym}(Q))$. 
Then by the definition of $\progr{P}{E}$ it must be that $l_\phi$ is in $(R\cup U)$ as defined in Def.~\ref{defn:rpmep-progression}, i.e., it is a fluent to be deleted due to a conditional effect of $E$ or it is an uncertain firing fluent. Then we have that $(\gamma, \neg \acting \phi)\in E$ and $\acting\phi\in P$, with $\gamma\in P$, or $(\gamma, \acting \negkb{\phi})\in E$ and $\negkb{\gamma} \not\in P$. Then by the definition of $\negeffs{E}$ in Def.~\ref{def:encoding}, rules (\ref{eq:uncertainty-closure})-(\ref{eq:uncertainty-ce}) and the definition of $result$, it must be the case that $l_\phi$ is not in $result(\invertedbelief{\actingsym}(P)\cup \ninvertedbelief{\actingsym}(P),E_o)$ either. 

Assume now $l_\phi$ is in  $\closure(\invertedbelief{\actingsym}(Q)\cup \ninvertedbelief{\actingsym}(Q))$ and $\closure(\invertedbelief{\actingsym}(P)\cup \ninvertedbelief{\actingsym}(P))$ but not in $result(\invertedbelief{\actingsym}(P)\cup \ninvertedbelief{\actingsym}(P),E_o)$. 
This means that  $\condeff{\cond}{\lit_\phi}\in\negeffs{E}(s)$ for some $\cond$, with $s=\closure(\invertedbelief{\actingsym}(P)\cup \ninvertedbelief{\actingsym}(P))$. 
Then, by the definition of $\negeffs{E}$ in Def.~\ref{def:encoding}, it must be the case that $(\gamma, \neg \acting \phi) \in \mathit{E}$. As a consequence, $\phi$ is in the set $R$ as defined in Def.~\ref{defn:rpmep-progression}, which implies that it cannot be in   $\closure(\invertedbelief{\actingsym}(Q)\cup \ninvertedbelief{\actingsym}(Q))$, by definition of  $Q=\progr{P}{E}$.

The cases in which $l_\phi$ is not in  $\closure(\invertedbelief{\actingsym}(Q)\cup \ninvertedbelief{\actingsym}(Q))$ nor in  $\closure(\invertedbelief{\actingsym}(P)\cup \ninvertedbelief{\actingsym}(P))$ but  in $result(\invertedbelief{\actingsym}(P)\cup \ninvertedbelief{\actingsym}(P),E_o)$, or conversely not in $\closure(\invertedbelief{\actingsym}(P)\cup \ninvertedbelief{\actingsym}(P))$ nor in  $result(\invertedbelief{\actingsym}(P)\cup \ninvertedbelief{\actingsym}(P),E_o)$ but in $\closure(\invertedbelief{\actingsym}(Q)\cup \ninvertedbelief{\actingsym}(Q))$ can be easily ruled out by the definition of $\poseffs{E}$ in  Def.~\ref{def:encoding} and the closure rule (\ref{eq:closure}), as $\closure(P) = \{\literala \mid P \vDash \literala\}$. Indeed, all and only the positive conditional effect of a \epplan action are encoded as positive conditional effects in the FOND planning problem. 
\end{proof}

With these extra conditional effects, we have a faithful encoding of the original \epplan problem. In other words, a policy $\alpha$ will be found\footnote{This assumes a sound and complete planning algorithm for the encoded problem.} for initial state $\init$ and goal $\goal$: $\init \models [\alpha]\goal$.



\begin{theorem}
\label{thm:solution-correspondence}
Our encoding is sound and complete with respect to \epplan.
\end{theorem}

\begin{proof}
The proof is a straightforward extension of the proof for Theorem~\ref{thm:belief-update} and the fact that Definition~\ref{def:encoding} (together with the rules for ancillary conditional effects) faithfully encode $\progr{P}{E}$. Any policy that is derived using a sound and complete planner on the encoded problem will thus generate a sound and complete policy.
\end{proof}

\section{Extensions}
\label{sec:extensions}

Given the core compilation, there are a variety of extensions we have explored to improve the task of domain modeling. Here, we present some of the key ones that have had the largest impact.

\subsection{Conditioned Mutual Awareness}
\label{sec:ma-cond}

Our specification of a \epplan problem and the subsequent encoding into classical planning allow us to specify a rich set of actions. Unlike traditional approaches that compile purely ontic action theories into ones that deal with belief (e.g., the work on conformant planning by \citeasnoun{PalaciosG09}), we allow for arbitrary conditional effects that include nested belief both as conditions and as effects.


While expressive, manually encoding effects with nested belief can be involved due to the cascading of ancillary conditional effects. Here, we extend the scope of ancillary conditional effects to safely capture a common phenomenon in planning with nested belief: mutual awareness of the effects of actions.

\begin{example}
In our running example, if an agent enters a room, then we realize this as an effect: e.g., $\condeff{\la \emptyset, \emptyset \ra}{at\_1\_loc1} \in \poseffs{}$. In many applications, other agents may also be aware of this: e.g., $\condeff{\la \emptyset, \emptyset \ra}{\belief{2} at\_1\_loc1} \in \poseffs{}$. Perhaps we wish to predicate this effect on the second agent believing that it is also in this  room: e.g., $\condeff{\la \set{\belief{2} at\_2\_loc1}, \emptyset \ra}{\belief{2} at\_1\_loc1} \in \poseffs{}$. It is this kind of behaviour of conditioned mutual awareness that we would like to capture in a controlled and automated manner.
\end{example}

By appealing to ancillary conditional effects, we will create new effects from existing ones. We have already demonstrated the ancillary conditional effects required for a faithful encoding to adhere to the axioms and state constraints we expect from our agent. We extend this idea here to capture the appealing property of \emph{conditioned mutual awareness}.
%




\begin{definition}{Conditioned Mutual Awareness}
An \epplan planning action $a \in \A$ is a tuple \(\la \pre_a, \macond{a}{i}, \Eff_a \ra\), in which $\pre_a$ and $\Eff_a$ are the same as in Definition~\ref{defn:rp-mep-planning-action}, and $\macond{a}{i} \in \fluents$ is the condition for agent $i$ to be aware of the effects of action $a$. Note that $\macond{a}{i}$ can be a unique fluent that is either always believed or never believed by a given agent.
\end{definition}

Intuitively, agent $i$ is aware of every conditional effect of $a$ only when agent $i$ believes $\macond{a}{i}$. 

For a given set of fluents $T$, we define the shorthand $\belief{i} T = \set{\belief{i} l \st l \in T}$ and $\neg \belief{i} T = \set{\neg \belief{i} l \st l \in T}$ and model conditioned mutual awareness through the following two encoding rules for every agent $i \in \agents$ to derive new conditional effects:
\begin{align}
\condeff{\la \condp, \condn \ra}{\lit} \in \poseffs{a} &\Rightarrow \notag\\
\condeff{\la \belief{i} \condp \cup \neg \belief{i} \condn &\cup \set{\belief{i} \mm{1} \macond{a}{i} \mm{1}}, \emptyset \ra}{\belief{i} \lit} \in \poseffs{a} 
\stepcounter{equation}\tag{\theequation}\label{eq:condition-awareness-positive} \\
\condeff{\la \condp, \condn \ra}{\lit} \in \negeffs{a} &\Rightarrow \notag\\
\condeff{\la \belief{i} \condp \cup \neg \belief{i} \condn &\cup \set{\belief{i} \mm{1} \macond{a}{i} \mm{1}}, \emptyset \ra}{\neg \belief{i} \lit} \in \poseffs{a}
\label{eq:condition-awareness-negative}
\end{align}

Note that each form of ancillary conditional effect adds a new positive conditional effect. In the positive case, we  believe that the agent $i$ has a new belief $\belief{i} \lit$ if we believe that agent $i$ had the prerequisite belief for the effect to fire. In the negative case, we would believe that the agent no longer holds the belief, but because we take a perspectival view, it is encoded as a positive conditional effect -- i.e., we would believe $\neg \belief{i} \lit$.
For instance, the ancillary conditional effect 
%
%
%
from our working example says that we should no longer believe the negation of agent 1's secret if we do not believe agent 1 is untrustworthy (see Example \ref{ex:trustworthy}), which would create the following ancillary conditional effect: 
\begin{align*}
\condeff{\la \emptyset, \set{\neg t_1} \ra}{\neg s_1} \in \negeffs{} & \Rightarrow \\
\condeff{\la \set{\neg \belief{2} \neg t_1}, \emptyset \ra}{& \mm{1} \neg \belief{2} \neg s_1} \in \poseffs{}\!\!. 
\end{align*}

%
%

We restrict the application of the above rules by applying them only if the following two conditions are met: (1) every RML in the newly created effect has a nested depth smaller than or equal to our bound $d$; and (2) if we are applying the above rule for agent $i$ to a conditional effect $\condeff{\cond}{\lit} \in \negeffs{o}$, then $\lit \notin \set{\belief{i} \lit', \neg \belief{i} \lit'}$. The first restriction bounds the number of conditional effects while the second prevents unwanted outcomes from introspection.
To see why this exception is required, consider the example of a pair of conditional effects %
%
for an action where we discover agent 1 may or may not believe $s_2$ (i.e., we should forget any belief about what agent 1 believes regarding $s_2$). Omitting $\macond{o}{1}$ for clarity, we have the following \emph{negative} conditional effects:
\begin{align*}
\condeff{\la \emptyset, \emptyset \ra}{\neg \belief{1} s_2} \hspace{3mm} \condeff{\la \emptyset, \emptyset \ra}{\belief{1} s_2}
\end{align*}

If we were to apply the above rules with agent $1$, we would add two \emph{positive} ancillary conditional effects:
\begin{align*}
\condeff{\la \emptyset, \emptyset \ra}{\neg \belief{1} \neg \belief{1} s_2} \hspace{3mm} \condeff{\la \emptyset, \emptyset \ra}{\neg \belief{1} \belief{1} s_2}
\end{align*}
which subsequently would simplify to the following conditional effects (given that we combine  successive modalities of the same agent index under \kd):
\begin{align*}
\condeff{\la \emptyset, \emptyset \ra}{\belief{1} s_2} \hspace{3mm} \condeff{\la \emptyset, \emptyset \ra}{\neg \belief{1} s_2}
\end{align*}
Thus, the resulting effects would indicate that 
the agent reaches an inconsistency with its own belief. To avoid this issue, we apply rule (\ref{eq:condition-awareness-negative}) only when the effect is not a belief (negative or positive) of the corresponding agent.

Because we can assume that conditioned mutual awareness is given and computed in the original \epplan specification, Theorem \ref{thm:solution-correspondence} continues to hold. \\ \smallskip 

\vspace{-1em}
\subsection{Always Known Fluents}
\label{sec:always-known}

Some of the time, some things are universal. This is the philosophy behind our powerful second extension: \textit{always known fluents} (or AK fluents for short).
These specially designated fluents are such that \textit{every} agent \textit{always} knows their value (true or false), and it is common knowledge among all agents that this is the case.
Having this capability allows us to effectively isolate the epistemic fragment of a planning problem from the rest of the combinatorics involved -- AK fluents behave just as regular fluents in planning and require no special treatment to capture nested belief surrounding them. As a common example, static fluents describing the layout of a map are always commonly known among the agents, and need not have nested belief associated with them.

Frequently, syntactic sugar for a planning language simplifies the formulation of a model but does not result in an improved efficiency for solving. This is not the case with AK fluents. Not only does it simplify the modeling -- allowing the domain designer to focus on only those aspects which should be epistemic in isolation -- it further improves the planner's capability of solving the problem because we do not require encoding new fluents for the nested belief of AK fluents, nor do we require additional effects to handle the nested beliefs. There is no impact on the space of valid plans, but without the AK treatment, the classical planner must maintain the nested belief of all agents for these fluents which is largely redundant.
We show how AK fluents are incorporated to seamlessly work with the encoding presented in Section~\ref{sec:encoding}, and the following updated version of Definition~\ref{def:encoding} highlights the key differences in \textbf{bold} font.

\begin{definition}{Encoding of \epplan with AK Fluents}
Given a \epplan problem, \(  \la \P, \A, \agents, \init, \goal \ra \) where $\P = \Preg \cup \Pak$ is composed of regular fluents $\Preg$ and AK fluents $\Pak$ (where $\Preg \cap \Pak = \emptyset$), and a bound $d$ on the depth of nested belief we wish to consider, we define the classical encoding as the tuple $\la \fluents, I, G, \operators \ra$ such that:
\begin{align*}
\bm{AK(KB)} &\bm{\myisdef \set{l_\phi \mid \phi \in (KB \cap \Pak)}} \\
\bm{\overline{AK}(KB)} &\bm{\myisdef \set{l_\phi \mid \neg\phi \in KB \textrm{ and } \phi \in \Pak)}} \\
F &\myisdef \set{l_\phi \st \phi \in \allrmls(\Preg)} \bm{\cup AK(\P)} \\
I &\myisdef \invertedbelief{\actingsym}(\closure{(\init)}) \bm{\cup AK(\closure{(\init)})} \\
G &\myisdef \invertedbelief{\actingsym}(\goal) \bm{\cup AK(\goal)}
\end{align*}
\noindent and for every action $\la \pre_a, \Eff_a \ra$ in $\A$, we have a corresponding operator $\la \pre_o, \Eff_o \ra$ in $\operators$ such that:
\begin{align*}
\pre_o \myisdef &\invertedbelief{\actingsym}(\pre_a) \bm{\cup AK(\pre_a)}
\end{align*}
and for every set of conditional effects $E \in \Eff_a$:
\begin{align*}
\poseffs{E} \myisdef &\set{\condeff{\la \invertedbelief{\actingsym}(\gamma) \bm{\cup AK(\gamma)}, \overline{\ninvertedbelief{\actingsym}(\gamma)} \bm{\cup \overline{AK}(\gamma)}\ra}{\lit_\phi} \st (\gamma, \acting \phi) \in \mathit{E}} \mm{2} \cup \\
&\set{\condeff{\la \bm{AK(\gamma)}, \bm{\overline{AK}(\gamma)}\ra}{\lit_\phi} \st \bm{(\gamma, \phi) \in \mathit{E}} \mm{1} \mathrm{\textbf{and}} \mm{1} \bm{\phi \in \Pak}} \\
\negeffs{E} \myisdef &\set{\condeff{\la \invertedbelief{\actingsym}(\gamma) \bm{\cup AK(\gamma)}, \overline{\ninvertedbelief{\actingsym}(\gamma)} \bm{\cup \overline{AK}(\gamma)}\ra}{\lit_\phi} \st (\gamma, \neg \acting \phi) \in \mathit{E}} \mm{2} \cup \\
&\set{\condeff{\la \bm{AK(\gamma)}, \bm{\overline{AK}(\gamma)}\ra}{\lit_\phi} \st \bm{(\gamma, \neg \phi) \in \mathit{E}} \mm{1} \mathrm{\textbf{and}} \mm{1} \bm{\phi \in \Pak}}
\end{align*}
\end{definition}

Note that the AK fluents can be used as conditions for regular effects, but if they are changed as part of any effect then only AK fluents may appear as conditions. We make this assumption as part of the AK extension, as otherwise uncertainty could propagate from non-AK fluents to AK fluents. With the restriction in place, every agent will know the true value of every fluent in $\Pak$ (and all have common knowledge that this is the case).

Generally speaking, we maintain the property that AK fluents are treated as if they are common knowledge, and use the absence of any particular AK fluent $\phi \in \Pak$ in the state to represent the fact that $\phi$ is commonly known among the agents to be false.

Aside from the core encoding, we also must consider how the fluents are treated in the definition of ancillary conditional effects. Depending on the type of effect, and where the AK fluents exist, we have the following:

\begin{itemize}
    
    \item Ancillary effects (\ref{eq:remove-negation})-(\ref{eq:uncertainty-closure}) are applied as normal if the literal being changed is from $\Preg$, and not applied at all if it is from $\Pak$.
    
    \item For ancillary effect (\ref{eq:uncertainty-ce}), if the original effect is adding a $\Pak$ literal, then it does not need to fire (recall that the negation of a literal from $\Pak$ will never appear in the state, as the absence of the positive literal means that it's commonly known to be false). If, on the other hand, the original effect is adding a $\Preg$ literal, then the ancillary effect remains unchanged -- negated $\Pak$ literals may be placed in the $\condn$ set as a result, but this is benign as they will never appear in the agent's knowledge base.
    
    
    \item Finally, ancillary effects (\ref{eq:condition-awareness-positive}) and (\ref{eq:condition-awareness-negative}) are modified to treat the $\Pak$ literals uniquely, and can only be used for effects that add literals from $\Preg$. The following is an updated version of (\ref{eq:condition-awareness-positive}), and (\ref{eq:condition-awareness-negative}) is analogous:
    \begin{align*}
        ak(\cond) &\myisdef \set{\lit_\phi \st \lit_\phi \in \cond \mm{1} \mathrm{and} \mm{1} \phi \in \Pak}\\
        \belief{i} \cond &\myisdef \set{\belief{i} \lit_\phi \st \lit_\phi \in \cond \mm{1} \mathrm{and} \mm{1} \phi \in \allrmls(\Preg)}
    \end{align*}
    \begin{align*}
        \condeff{\la \condp, \condn \ra}{\lit} \in \poseffs{o} &\Rightarrow \notag\\
        \condeff{\la \belief{i} \condp \cup \neg \belief{i} \condn \cup \mm{1} &\belief{i} \set{\macond{o}{i}} \bm{\cup ak(\set{\macond{o}{i}})} \bm{\cup ak(\condp)}, \bm{ak(\condn)} \ra}{\belief{i} \lit} \in \poseffs{o} 
    \end{align*}
    Note that the condition for mutual awareness may be from $\Preg$ or $\Pak$.

\end{itemize}

It is worth reiterating the prevailing assumption regarding the literals from $\Pak$: \textit{if the positive literal is not in the agent's knowledge base, then it is commonly known to be false}. As a result, the above restrictions on the ancillary effects will never add the negation of a literal from $\Pak$ as an effect.

We do not detail a formal proof analogous to that of Theorem \ref{thm:solution-correspondence}, but generally the soundness and correctness can be seen as a direct result of treating the $\Pak$ fluents as common knowledge among all of the agents, and the modifications listed above simply following that assumption through proper bookkeeping.

\section{Action Examples}
\label{sec:examples}

\newcommand{\emptycond}[1]{\condeff{\la \emptyset, \emptyset \ra}{#1}}

To give a better sense of what is possible with our epistemic planning framework, we detail a few commonly used action types in the context of  nested belief. This list is by no means exhaustive, but simply serves to illustrate the modeling possibilities.

\subsection{Partially and Fully Observable Ontic Actions}



Consider a simple ontic action to turn the lights on, which can be captured as follows, which has the positive effect of enabling  $\mathit{lights\_on}$, and the negative effect of enabling  \(\mathit{lights\_off} \): 

\begin{align*}
    \pre = \set{\mathit{lights\_off}} \\
    \macond{}{i} = \top \\
    \Eff = \set{\la \poseffs{}, \negeffs{} \ra} \\
    \emptycond{\mathit{lights\_on}} \in \poseffs{} \\
    \emptycond{\mathit{lights\_off}} \in \negeffs{}
\end{align*}

Suppose instead we were interested in a general flip action. If we assume that  $\mathit{lights\_on}$ is an \textsc{AK} fluent (otherwise, we would need negative preconditions), then we could use: 


\begin{align*}
    \pre = \emptyset \\
    \macond{}{i} = \top \\
    \Eff = \set{\la \poseffs{}, \negeffs{} \ra} \\
    \condeff{\la \set{\mathit{lights\_on}}, \emptyset \ra}{\mathit{lights\_on}} \in \negeffs{} \\
    \condeff{\la \emptyset, \set{\mathit{lights\_on}} \ra}{\mathit{lights\_on}} \in \poseffs{}
\end{align*}

Note that both of these examples consider a fully observable environment, and so to capture partial observability, we could use: 


\begin{align*}
    \pre = \set{\mathit{lights\_off}} \\
    \macond{}{i} = \mathit{in\_room\_i} \\
    \Eff = \set{\la \poseffs{}, \negeffs{} \ra} \\
    \emptycond{\mathit{lights\_on}} \in \poseffs{} \\
    \emptycond{\mathit{lights\_off}} \in \negeffs{}
\end{align*}
in which the condition on mutual awareness is that the agent is in the room: $\mu_i = in\_room\_i$.

\subsection{Public and Semi-private Announcements}

Let us now consider the case of announcements. Interestingly, we capture both truthful and untruthful announcements in a simple fashion. For example, the following captures a truthful public announcement that the door is open: 



\begin{align*}
    \pre = \set{\mathit{door\_open}} \\
    \macond{}{i} = \top \\
    \Eff = \set{\la \poseffs{}, \emptyset \ra} \\
    \forall i \in \agents \setminus \set{\actingsym}, \emptycond{\belief{i} \mathit{door\_open}} \in \poseffs{}
\end{align*}

If we now remove the precondition (that is, set it to the empty set), the announcement could be an untruthful one. 


Suppose we are now interested in making a a semi-private truthful announcement, in the sense that the announcement is heard by all agents present in a certain room, we could use: 
\begin{align*}
    \pre = \set{\mathit{door\_open}} \\
    \macond{}{i} = \mathit{in\_room\_i} \\
    \Eff = \set{\la \poseffs{}, \emptyset \ra} \\
    \forall i \in \agents, \condeff{\la\set{\mathit{in\_room\_i}}, \emptyset\ra}{\belief{i} \mathit{door\_open}} \in \poseffs{}
\end{align*}

\subsection{Yes/No Questions}

Suppose we are interested in inquiring whether \textit{i} believes the lights are on. Such an action is captured as follows: 
\begin{align*}
    \pre = \emptyset \\
    \macond{}{i} = \top \\
    \Eff = \set{\la \poseffs{1}, \emptyset \ra, \la \poseffs{2}, \emptyset \ra} \\
    \emptycond{\belief{i} \mathit{lights\_on}} \in \poseffs{1} \\
    \emptycond{\belief{i} \neg \mathit{lights\_on}} \in \poseffs{2}
\end{align*}

Such interactions have appeared in works such as  \cite{coin-journal-teamwork} on teamwork formation, where negotiation between agents is conducted through a series of yes/no questions similar to the one presented here.


\section{Evaluation}
\label{sec:evaluation}

In this section, we present an evaluation of our ideas over several planning domains --- some inspired from previous literature and others in which our planner has been used to generate solutions. We evaluate over a series of parameters, including number of agents and maximum depth of a restricted modal literal. The vast majority of domains explore settings where inconsistent or incorrect belief play a role -- a key strength of our planner compared to many epistemic reasoning systems available.

We implemented the scheme above to convert a \epplan planning problem into a classical planning problem, which can be subsequently solved by any planner capable of handling negative preconditions and conditional effects. The compiler consumes a custom format for the \epplan problems and can either simulate the execution of a given action sequence or invoke a classical planner built using a configuration of the LAPKT planning library \cite{lapkt}. The source code, benchmarks, and demo of the compilation process can be found online at:
\begin{center}
\url{http://pdkb.haz.ca/}
\end{center}

Throughout this section, we will use the following notation:
\begin{itemize}
    \item $\agents$: the set of the agents in the problem
    \item $d$: the maximum depth of nested reasoning
    \item $F$: the fluents in the encoding
    \item $\opseq$: the computed sequential plan
\end{itemize}

Further, to emphasize one key strength of our choice to build on top of classical planners, we delineate the difference between the planner originally used when we introduced the system \cite{muise2015planning} with the latest incarnation of the same planner. In some instances, we find a substantial improvement in computation efficiency, and this is directly attributed to using classical planning as a black-box technology: as the field progresses, so does the strength of our approach.

\subsection{Thief}

The \emph{Thief} \cite{LowePW11} problem is one of deception. A simplified version of a computer game, the thief agent must steal an item while avoiding being detected by guards. The thief agent can use actions to misdirect the guards' `attention'. The actions in the domain model events in which the one party can make a noise and the other party notices; one can see the other from behind, learning of their location; and both can see each other face-to-face, simultaneously learning of each others' locations. In the problem instances, the locations of each is initially unknown.

We have verified the model of the pre-existing \emph{Thief} problem, and all of the existing queries considered in the previous literature posed to demonstrate the need for nested reasoning (e.g., those found in \cite{LowePW11}) are trivially solved in a fraction of a second.

\subsection{Corridor}

The \emph{Corridor} problem involves agents that can walk back and forth between rooms of a corridor, and state information that they believe to be true within earshot of the neighbouring rooms \cite{kominisbeliefs}. Specifically, an agent broadcasting proposition $q$ in a room will then be believed by all agents in the same room as well as neighbouring rooms. As the setting is for a framework of common knowledge, the belief of proposition $q$ is common among all agents that hear the statement, and the position of all agents is universally known as well.

This domain was modified to allow for the broadcasting agent to spread false belief (i.e., to lie about the secret). All those agents within earshot of the broadcast will adopt whatever is announced, and the goals contain a mix of agent belief over what is believed to be correct / incorrect.

We varied some of the discussed parameters and report on the results in Table \ref{tbl:results-corridor} (the first Corridor problem corresponds to the one presented by \citeasnoun{kominisbeliefs} with the exception of the ability to lie). As the results show, as the depth and number of agents increase, the compilation time increases exponentially, while the planning time is standard for a classical planning problem of this scale. There was little difference between the old and new planners, and only one column for plan size is included as the two planners coincided in this regard for every problem.

\begin{table}[ht]
	\centering
	\begin{tabular}{ccrrrrrr}
		\toprule
		\multirow{2}{*}{$|\agents|$} & \multirow{2}{*}{$d$} & \multirow{2}{*}{$|F|$} & \multirow{2}{*}{$|\opseq|$} & \multicolumn{4}{c}{\textbf{Time (s)}} \\
	  &  &  &  & \multicolumn{1}{l}{\textbf{Solve$_{old}$}} & \multicolumn{1}{l}{\textbf{Solve}} & \multicolumn{1}{l}{\textbf{Compile}} & \textbf{Total} \\
		\midrule
%
%
%
%
3 & 1 & 54 & 8 & 0.01 & 0.03 & 0.10 & 0.13 \\
5 & 1 & 62 & 8 & 0.01 & 0.03 & 0.14 & 0.17 \\
7 & 1 & 70 & 8 & 0.01 & 0.03 & 0.18 & 0.20 \\
3 & 3 & 558 & 8 & 0.02 & 0.04 & 0.78 & 0.82 \\
5 & 3 & 2262 & 8 & 0.10 & 0.13 & 4.64 & 4.77 \\
7 & 3 & 5950 & 8 & 0.60 & 0.68 & 15.65 & 16.33 \\
3 & 5 & 18702 & 8 & 3.34 & 3.33 & 55.69 & 59.02 \\
5 & 5 & 222262 & MO & MO & MO & 1776.10 & MO \\
7 & 5 & MO & MO & MO & MO & MO & MO \\
%
 %
 %
 %
   		\bottomrule
	\end{tabular}
	\caption{Results for encoding and planning time for the Corridor problem.}
	\label{tbl:results-corridor}
\end{table}

At the extreme end, we found the limit for both what the planning procedure and our preprocessing approach can accomplish. With 5 agents and depth 5, we have over 200k fluents created. The compilation to classical planning completes (in just under half an hour), but the planner runs out of memory trying to solve the problem. Scaling further (with 7 agents), we are not even able to compile the problem.
While it is useful to examine the limits of our approach, we should emphasize that the majority of interesting use cases we have found for planning with nested belief is restricted to depth 1-2.

\subsection{Grapevine}

As a more challenging test-bed, we modelled a setting that combines the \emph{Corridor} problem \cite{kominisbeliefs} and the classic \emph{Gossip} problem \cite{Entringer1979353}. In the new problem, \emph{Grapevine}, there are three rooms with all agents starting in the first. Every agent believes their own secret to begin with, and the agents can either move between rooms or broadcast a secret they believe (or its negation if they wish to lie). Movement is always observed by all, but through the use of conditioned mutual awareness the sharing of a secret is only observed by those in the same room.
Unlike the corridor domain above, agents adopt a belief only if they do not already have an assumption about it. That is, we modelled the notion that an agent ``cannot change their mind''. This problem allows us to pose a variety of interesting goals ranging from private communication (similar to the Corridor problem) to goals of misconception/deception in the agent's belief (e.g., $G = \set{\belief{a} s_b, \belief{b} \neg \belief{a} s_b}$).
 
 \begin{table}[ht]
 	\centering
 	\begin{tabular}{cccrrrrrrr}
 		\toprule
 		\multirow{2}{*}{$|\agents|$} & \multirow{2}{*}{$|g|$} & \multirow{2}{*}{$d$} & \multirow{2}{*}{$|F|$} & \multirow{2}{*}{$|\opseq|_{old}$} & \multirow{2}{*}{$|\opseq|$} & \multicolumn{4}{c}{\textbf{Time (s)}} \\
 		 &  &  &  &  &  & \multicolumn{1}{l}{\textbf{Solve$_{old}$}} & \multicolumn{1}{l}{\textbf{Solve}} & \multicolumn{1}{l}{\textbf{Compile}} & \textbf{Total} \\
 		\midrule
%
%
%
4 & 2 & 1 & 116 & 4 & 4 & 0.07 & 0.08 & 0.81 & 0.89 \\
4 & 4 & 1 & 116 & 6 & 6 & 0.10 & 0.07 & 0.83 & 0.91 \\
4 & 8 & 1 & 116 & 16 & 12 & 0.06 & 0.08 & 0.84 & 0.93 \\
4 & 2 & 2 & 628 & 6 & 5 & 2.31 & 1.39 & 13.37 & 14.75 \\
4 & 4 & 2 & 628 & 7 & 7 & 12.13 & 1.37 & 13.24 & 14.61 \\
4 & 8 & 2 & 628 & 22 & 27 & 216.15 & 2.35 & 13.02 & 15.37 \\
8 & 2 & 1 & 340 & 4 & 4 & 1.02 & 0.65 & 5.49 & 6.14 \\
8 & 4 & 1 & 340 & 9 & 11 & 2.45 & 1.95 & 5.47 & 7.42 \\
8 & 8 & 1 & 340 & 16 & 16 & 2.36 & 0.83 & 5.52 & 6.35 \\
8 & 2 & 2 & 4436 & 6 & 5 & 350.23 & 309.79 & 253.69 & 563.48 \\
8 & 4 & 2 & 4436 & 12 & 9 & 650.78 & 303.81 & 260.27 & 564.07 \\
8 & 8 & 2 & 4436 & TO & 17 & TO & 337.48 & 254.20 & 591.69 \\
%
%
 		\bottomrule
 	\end{tabular}
 	\caption{Results for encoding and planning time for the Grapevine problem.}
 	\label{tbl:results-grapevine}
 \end{table}

Table \ref{tbl:results-grapevine} shows the results of our system in this more involved setting. We additionally report the size of the goal posed to the planner as $|g|$. Note that the results on plan length between the old and new planner are roughly similar, but there are a number of problems were the new planner finds the solution in far less time (e.g., two orders of magnitude improvement on the 6th problem).
 


We discuss the most related epistemic planners in Section \ref{sec:related}, and more importantly why they cannot be used as a comparison on these problems. However, unlike the Corridor domain above and the Grid domain below, the Grapevine domain can be modelled in the language used by the newly introduced epistemic planner EFP2.0 \cite{FabianoBDP20}. Table \ref{tbl:results-efp} shows a comparison for the first 6 problems (those with 8 agents are not solvable due to memory violation). The resource limits for this evaluation were 1 hour and 32Gb of memory using a stronger machine than the other evaluations.\footnote{We thank the lead author of EFP2.0, Francesco Fabiano, for help running this evaluation.} The plan lengths were equivalent, and the discrepancy on this measure for lines 4-5 stem from the additional ``initialize'' action that is used in the RP-MEP encoding.

\begin{table}[ht]
 \centering
 \begin{tabular}{ccc|cc|rrr}
 	\toprule
 	\multirow{2}{*}{$|\agents|$} & \multirow{2}{*}{$|g|$} & \multirow{2}{*}{$d$} & \multicolumn{2}{c|}{$|\opseq|$} & \multicolumn{3}{c}{Time (s)} \\
 	 &  &  & RPMEP & EFP2 & RPMEP & \multicolumn{1}{l}{EFP2} & \multicolumn{1}{l}{EFP2$_{simp}$} \\
 	\midrule
 	4 & 2 & 1 & 4 & 4 & 0.46 & 39.40 & 0.67 \\
 	4 & 4 & 1 & 6 & 6 & 0.47 & 2698.30 & 100.22 \\
 	4 & 8 & 1 & 12 & TO & 0.46 & TO & - \\
 	4 & 2 & 2 & 5 & 4 & 9.28 & 48.43 & 0.91 \\
 	4 & 4 & 2 & 7 & 6 & 9.26 & 3413.02 & 14.37 \\
 	4 & 8 & 2 & 27 & TO & 9.76 & TO & - \\
 	\bottomrule
 \end{tabular}
 \caption{Results for encoding and planning time for EFP2.0 on the Grapevine problem.}
 \label{tbl:results-efp}
\end{table}
 
The column for ``simplified'' demonstrates the solve time if the problem is simplified so fluents irrelevant to the goal are manually pruned. This is a fairly straightforward process to approximate, and captures the performance if this simple preprocessing were done automatically.

We find that as the general plan length grows, as driven by the complexity of the goals and nesting used to achieving them, the scalability of EFP2.0 suffers. In some sense, this is to be expected given the richness of the formalism handled by EFP2.0: complex formulae (including those with disjunction) can be used which is not the case for RP-MEP. Further, the representation is built using accessibility relations which means that scaling of nested depth has little to no impact on the performance.

This domain serves as a prime example of the orthogonal nature of RP-MEP and other epistemic planners capable of handling false belief: the scalability and efficiency seen in RP-MEP comes from its ability to reason over complex action theories in large state spaces, while its limitation stems from the depth of nesting and type of formulae that is permitted.

\subsection{Selective Communication}

This is a collection of domains and problems in the area of \emph{selective communication} \cite{wu2011online}. In a multi-agent system, selective communication is the task of deciding when and what to communicate between agents to improve the outcomes of a collective task. For example, in a disaster response scenario, the environment will be partially-observable and initially, much of it will be unknown. As individual agents survey the area, they will obtain knowledge about the domain and can communicate this information to update  other agents. However, communicating all information can be non-optimal if communication comes at a cost; e.g.\ if some agents are human who can become easily overwhelmed, or if there is a cost of communicating, such as a risk of giving away one's location in an adversarial environment.  Using epistemic planning allows us to model communication as a natural action: communicating knowledge/belief to another agent updates their beliefs. We further consider the possibility of false beliefs propagating (e.g., through the malicious spread of misinformation or faulty communication).

The selective communication domains originally come from \citeasnoun{alshehri2017}, and define five different domains in which survivors must be located in a set of rooms and brought back to a medical area. The five domains are: (1) a simple scenario in which there is an unknown number of survivors whose location is initially unknown, and each survivor's location must be believed by at least one agent (distributed belief); (2) same as scenario 1 except that all agents must believe that all other agents believe the locations of all survivors (higher-order epistemic goal); (3) same as 1 except agents can broadcast belief and a single agent (the commander) needs to believe the location of all survivors, while all other agents can be ignorant of their location; (4) same as 2 except communication is one-to-one instead of using a broadcast; and (5) same as 1 except in a cluttered environment with unknown obstacles preventing certain plans, and so sharing belief about these plans can improve the agents' navigation.

The five different domains have several problems, by varying the number of agents and number of rooms to search, based on the well-known \emph{Blocks World For Teams} (BW4T) scenario \cite{johnson2009joint}.
In our evaluation, we take only the selective communication model outlined by \citeasnoun{alshehri2017}, omitting the two baselines of no communication and communicating all new information. Actions include moving between rooms, sensing survivors (epistemic), communicating with other agents (epistemic action), and transporting survivors back to the medical area.
All problems in this set make use of `always knowing' fluents (cf. Section~\ref{sec:always-known}) to model static parts of the problem, such as room IDs / locations.

\begin{table}[!ht]
	\centering
	\begin{tabular}{lcrrrrrrr}
		\toprule
		\multirow{2}{*}{\textbf{Scenario}} & \multirow{2}{*}{$|\agents|$} & \multirow{2}{*}{$|F|$} & \multirow{2}{*}{$|\opseq|_{old}$} & \multirow{2}{*}{$|\opseq|$} & \multicolumn{4}{c}{\textbf{Time (s)}} \\
		& & &  & & \multicolumn{1}{l}{\textbf{Solve$_{old}$}} & \multicolumn{1}{l}{\textbf{Solve}} & \multicolumn{1}{l}{\textbf{Comp.}} & \textbf{Total} \\
		\midrule
%
%
%
%
%
%
!epgoal: 1 & 3 & 594 & 15 & 15 & 0.03 & 0.05 & 0.70 & 0.75 \\
!epgoal: 2 & 3 & 864 & 21 & 21 & 0.05 & 0.07 & 1.11 & 1.18 \\
!epgoal: 3 & 4 & 720 & 14 & 14 & 0.05 & 0.07 & 1.10 & 1.16 \\
!epgoal: 4 & 4 & 1032 & 20 & 20 & 0.07 & 0.09 & 1.78 & 1.87 \\[2mm]
epgoal: 1 & 3 & 594 & 37 & 42 & 17.52 & 0.08 & 0.70 & 0.78 \\
epgoal: 2 & 3 & 864 & 46 & 53 & 10.88 & 0.13 & 1.14 & 1.27 \\
epgoal: 3 & 4 & 720 & 61 & 36 & 659.21 & 10.67 & 1.09 & 11.76 \\
epgoal: 4 & 4 & 1032 & 65 & 42 & 1607.87 & 18.29 & 1.77 & 20.06 \\[2mm]
broad: 1 & 3 & 600 & 19 & 19 & 0.08 & 0.12 & 0.59 & 0.71 \\
broad: 2 & 3 & 600 & 19 & 19 & 0.03 & 0.05 & 0.89 & 0.93 \\
broad: 3 & 3 & 870 & 25 & 25 & 0.07 & 0.06 & 1.23 & 1.29 \\
broad: 4 & 4 & 1040 & 25 & 25 & 0.06 & 0.09 & 1.84 & 1.93 \\[2mm]
!broad: 1 & 3 & 732 & 23 & 23 & 0.13 & 0.14 & 6.18 & 6.32 \\
!broad: 2 & 3 & 732 & 23 & 23 & 0.13 & 0.15 & 6.39 & 6.54 \\
!broad: 3 & 4 & 1264 & 30 & 30 & 0.61 & 0.60 & 25.54 & 26.14 \\
!broad: 4 & 4 & 1264 & 30 & 30 & 0.61 & 0.60 & 25.31 & 25.91 \\[2mm]
blocked: 1 & 3 & 1008 & 50 & 51 & 1.39 & 0.13 & 2.81 & 2.93 \\
blocked: 2 & 3 & 1416 & 58 & 55 & 0.23 & 0.23 & 5.12 & 5.35 \\
blocked: 3 & 4 & 1242 & 50 & 53 & 2.68 & 0.20 & 4.68 & 4.88 \\
blocked: 4 & 4 & 1728 & TO & 73 & TO & 0.59 & 8.57 & 9.16 \\
%
%
	    



		\bottomrule
	\end{tabular}
	\caption{Results for encoding and planning time for the Selective Communication problem.}
	\label{tbl:results-sc}
\end{table}

Table~\ref{tbl:results-sc} shows the results for these. The scenario labels \textbf{!epgoal}, \textbf{epgoal}, \textbf{broad}, \textbf{!broad}, and \textbf{blocked} respectively correspond to scenarios (1)-(5) described above.
As with the other domains, we see that the compilation takes a bulk of the time.
However, what this shows different to the Corridor and Grapevine problems is that the size of the compilation does not affect the classical planner. That is, given a compilation, long, non-trivial plans can be generated, in most case in under one second, despite the larger state space. We also find a substantial improvement in planner performance compared to the old system used.

\subsection{Hattari}
\label{sec:eval-hattari}

Hattari is a board game where partial information and nested belief plays a crucial role in the strategy and game mechanics \cite{hattari}. Players secretly look at a card with a unique number written on it, and then pass the card to the right for the next person to look at. After this initial phase, there is common knowledge shared among every successive pair of players, and the game unfolds through a mix of deduction and bluffing centered around the ability for individuals to reason about nested belief.

We modeled the mechanics of the game using the framework presented in this paper, and embedded the belief maintenance as part of a web application:\footnote{Source code for the site: \url{https://bitbucket.org/jekegren/hattari-project}}
\begin{center}
\url{http://hattari.haz.ca/}
\end{center}

\begin{figure}
    \centering
    \includegraphics[width=\linewidth]{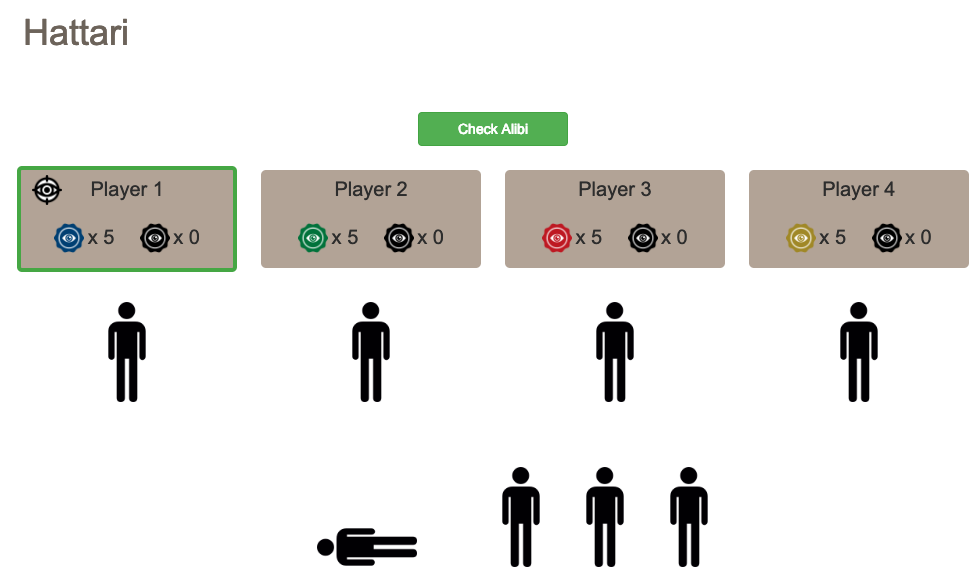}
    \caption{Screenshot of the online Hattari game}
    \label{fig:hattari}
\end{figure}

A screenshot of the online game can be seen in Figure \ref{fig:hattari}.

As planning is not a component to this scenario, we do not report on the evaluation time (which was negligible for the task of belief update). However, to get a sense of the encoding for the game, the following action demonstrates the perspectival model for agent $ag$ looking at a card $c$:


\begin{align*}
    \macond{}{i} = \top \\
    \Eff = \set{\la \poseffs{}, \emptyset \ra} \\
    \condeff{\la \set{\mathit{me_{ag}}}, \emptyset \ra}{\mathit{holding\_ag\_c}} \in \poseffs{} \\
    \condeff{\la \set{\mathit{holding\_ag\_c}}, \emptyset \ra}{\belief{ag}\mathit{holding\_ag\_c}} \in \poseffs{}
\end{align*}

We have the following properties of the action:
\begin{itemize}
    \item The precondition is not listed, as we are only doing belief maintenance and not planning (the game logic implemented elsewhere dictates what actions are possible to execute).
    \item The condition for mutual awareness is true, meaning that nested ancillary effects of the form in equation (\ref{eq:condition-awareness-positive}) will apply.
    \item $\mathit{me_{ag}}$ is a special fluent believed only by agent $\mathit{ag}$.
    \item The perspective of this action may not be from $\mathit{ag}$'s point of view.
\end{itemize}

The last point is of particular interest: when the perspective is from another agent, the first effect will not fire (as $\mathit{me_{ag}}$ is not believed), but the second one may. Other actions in the domain cause the cards to be rotated between players, and so a sequence such as $[\mathit{look\_ag\_c}, \mathit{pass\_cards}, \mathit{look\_ag2\_c}]$ from the perspective of $\mathit{ag}$ will cause the belief base to first contain $\mathit{holding\_ag\_c}$ from the first effect, and then $\mathit{holding\_ag2\_c}$ from the effects of the $\mathit{pass\_cards}$ action, and then $\belief{ag2} \mathit{holding\_ag2\_c}$ from the second effect of the third action.

This basic form of epistemic update demonstrates the scope of what our framework for epistemic planning can capture.

\subsection{Results Discussion}

The results in this section demonstrate that our approach can solve a wide variety of epistemic planning scenarios, from toy problems in the literature to larger-scale problems that contain longer plans. The trend shown in the results, though, is that the compilation time is the bottleneck in the approach; increasing exponentially in the number of agents and the depth of reasoning. The planning process is typically fast, and moving forward we hope to reduce the compilation time by only generating fluents and conditional effects that are relevant to achieving the goal.

\section{Related Work}
\label{sec:related}

\subsection{Comparison to Other Planners}
We contrast our work with the most closely related epistemic planners. In particular, two planners -- EFP and MEPK -- have the theoretical capability to handle the problems explored in Section \ref{sec:evaluation}. However, in practice, modelling and solving those settings are not feasible for a variety of reasons we detail here.\footnote{We would like to thank the authors of both systems for working with us to understand these core differences fully.}

\subsubsection{EFP}

A recent approach to epistemic planning is the EFP planner \cite{le-fab-son-pon-icaps18}. The authors present a pair of planners for generating multi-agent epistemic plans: EFP and PG-EFP.  Both are forward-search planners, differing with respect to their search mechanisms. EFP uses breadth-first search, while PG-EFP uses heuristic search based on a novel epistemic plan graph. These planners were originally inspired by and built on the ideas presented in this paper. 

Similar to RP-MEP, they syntactically restrict the epistemic formulae they can represent, however the encoding is different. The authors compare their planners to the \citet{huang2017general} planner and a version of RP-MEP that lacked the support for AK fluents, all on domains that required only notions of knowledge and not belief. Performance varies on different benchmarks. In general, RP-MEP demonstrates superior performance at shallow depth, but as d increases, EFP planners demonstrate superior performance. This is because the size of the RP-MEP problem grows exponentially with increased depth, which is not the case with the EFP planners.

Most crucially, while the language EFP is built on the $m\mathcal{A}^*$ action language, which is capable of representing incorrect beliefs of agents, 
the implementation of EFP focuses exclusively on nested knowledge. Core aspects of the benchmarks explored in Section \ref{sec:evaluation}, such as agents that share false information, is thus something we cannot model in EFP1.0 for solver comparison. We should emphasize, however, that this is a restriction only with the implementation of EFP1.0 and not with the $m\mathcal{A}^*$ language itself. Further, while RP-MEP is restricted to formulae that are disjunction-free, the EFP system has no such restriction.

Very recently, the EFP1.0 planner was extended to handle a richer class of epistemic problems in the EFP2.0 system \cite{FabianoBDP20}. The new system builds on the previous by defining an improved transition function, and includes a host of planner improvements such as duplicate state detection and reduced state size representation.
Unlike EFP1.0, the EFP2.0 implementation permits certain forms of inconsistent belief to be modeled. In particular, untruthful announcements are feasible, as long as it does not override prior belief of the agents that observe the announcement.

For the three main benchmarks explored in Section \ref{sec:evaluation}, only Grapevine follows this style of inconsistent belief.\footnote{See the link below in the discussion of the MEPK planner for the EFP2.0 model for the first Grapevine problem as an illustrative example.} The other two domains (Corridor and Grid) propagate incorrect belief in a manner that cannot yet be modelled by EFP2.0. On the other hand, the language handled by the EFP2.0 planner captures disjunction, and thus offers a far richer formalism in that regard. Further, as noted in the evaluation discussion, EFP2.0 is not impacted by the required depth of nested reasoning, whereas RP-MEP faces an exponential increase in the compiled representation size depending on the nested depth.

\subsubsection{MEPK}
\label{sec:mepk-comparison}

The most closely related planner to our work is MEPK \cite{huang2017general}. They take a different approach to epistemic planning by defining algorithms for belief revision and update over arbitrary \kdff formulae. Their approach exploits the use of \emph{alternating cover disjunctive formulas} (ACDF), first defined by \citet{hales2012refinement}. ACDF is used to represent the knowledge base and all other formulae in the planning problem, thus requiring a translation from arbitrary formula to ACDF. The length of an  ACDF formula is single exponentially bound by the length of the original formula. Checking entailment between two ACDF formulae is untractable, so \citet{huang2017general} introduce a stronger form of entailment, computable in polynomial time. This notion is then used to define polynomial-time belief revision and update operators. A standard search algorithm is then used to find plans, with the belief revision and update operators used to define progression. Similar to our notion of always known fluents, \citet{huang2017general} support static common knowledge, which are propositional formulae that never change and are common knowledge to all agents. Our always known fluents are more flexible in one regard, as they can be arbitrarily modified during execution, however they cannot contain disjunction. 

By supporting more expression formulae, \citet{huang2017general} are effectively trading efficiency for expressiveness. Their planner can handle arbitrary \kd formulae, while ours cannot, but their compilation in ACDF and search are more expensive than classical or FOND planning.

Unlike the majority of related work listed below, MEPK can handle true doxastic reasoning, and thus are able to handle notions such as agents that lie in communication. However, a full empirical comparison between the planners is not feasible due to the nature of our ancillary effects -- in particular those for uncertain firing and conditioned mutual awareness.

\ref{app:ancillary-domains} demonstrates the core ancillary effects, and the differences in representations used by RP-MEP and MEPK (PDKBDDL and EPDDL respectively). To further illustrate the incompatible nature, the following is the smallest problem in the grapevine domain for both languages (it also includes the format used by EFP2.0 as discussed above):

\begin{center}
    \url{http://editor.planning.domains/#read_session=UusNZTzIs3}
\end{center}

The PDKBDDL representation is under 100 lines in total, while the EPDDL is a semi-automated conversion from the compiled form of the PDKBDDL resulting in over 12,000 lines for the EPDDL encoding. The reason the EPDDL representation is so large is a consequence of having to compile all of the epistemic inferences into the model similar to RP-MEP. This is necessary since some key inference steps are not handled natively by the latest version of MEPK, and thus the full range of ramifications must be provided in the encoding.

RP-MEP takes under a second to solve the problem from start to finish including the preprocessing (see Section \ref{sec:evaluation} for the details), while MEPK runs out of memory in a matter of seconds (both with and without heuristics enabled). The expected plan length is 4, and the classical planner finds this in 0.06 seconds.

It should be emphasized that \textit{this is by no means a fair comparison}. It is essentially asking MEPK to reason about a classical encoding with all of the advanced doxastic reasoning compiled into the domain; a comparison between the expressive MEPK planner and a state-of-the-art classical planner. Many of the ancillary effects we use are handled natively by MEPK, but must be expanded due to their interaction with the other ancillary effects as a chain of reasoning. We detail these further in \ref{app:ancillary-domains}.

Conversely, there is no natural re-encoding that avoids this explosion in representation. The main reason is that some of the automatic inferences made by RP-MEP -- namely the conditioned mutual awareness and uncertain firing effects -- are interleaved with the remaining inferences that both planners handle. Thus, any effect added by RP-MEP's compilation that requires a chain of reasoning with at least one ancillary effect unique to RP-MEP will not be captured by the native MEPK reasoning. It must be encoded by hand.

Finally, we should emphasize that there is a large space of models that can be captured by MEPK that cannot be readily modelled and solved by RP-MEP. In particular, any domain that requires reasoning over disjunctions is something that distinguishes the two planners from one another. Additionally, the manner with which we model nested belief means that the scalability of RP-MEP is largely restricted to small depths, while MEPK does not face such issues.

Ultimately, the two planners offer complementary strengths, and the problems readily modelled in both PDKBDDL and EPDDL fail to exhibit the true strengths of either RP-MEP or MEPK. Combining the strengths of both approaches is an extremely lucrative area for future work.

\subsection{Additional Related Work}

There is a variety of research related to the ideas we have presented, which we briefly summarize below. 

%

Reasoning about knowledge and belief has a long history in philosophy, but has been gaining increasing prominence in computer science and AI \cite{Fagin} since the work of Kripke, Hintikka, Prior, among others  \cite{DBLP:journals/jsyml/Kripke59,nla.cat-vn2250548,prior1967past}. In particular, at least since the eighties, the extension of epistemic logic to reason about actions has received considerable attention, in languages such as the situation calculus and propositional modal logic \cite{moore1985,Reiter,Ditmarsch}. 

%
Our approach builds on the DEL family of languages. Research into DEL \cite{Ditmarsch}, and more recently DEL planning (e.g., \cite{BolanderA11}), deals with how to reason about knowledge or belief in a setting with multiple agents, in the interest of achieving individual or joint goals.  Until recently, focus in this area has  primarily been on the logical foundations (e.g., semantic considerations) for updating an epistemic state  according to physical (ontic) and non-physical (epistemic) actions, as well as identifying the classes of restricted reasoning that are tractable from a computability  standpoint \cite{LowePW11}. A preliminary version of our article \cite{muise2015planning} was among the first  approaches to consider the implementation perspective on epistemic planning. In particular, we leveraged  intuitions from state-of-the-art approaches for automated planning with partial observability, discussed further below.
This necessitated some restrictions on the specification language. While the full language of DEL is, therefore, clearly more expressive than our approach in terms of the logical reasoning that an agent can achieve in theory, this expressiveness increases the computational complexity of the reasoning. In particular, the practical synthesis of DEL plans remains a challenging problem. See, for example, \cite{engesser2017cooperative}  on notable progress on this front where the planner manipulates DEL models directly. We refer interested readers to international workshops such as \cite{baral_et_al:DR:2017:8285} for  recent advancements, as well as disparate communities  that approach epistemic planning from the perspective of game-theoretic strategies, linguistics, and so on.  (For example, outside of the reasoning about actions community, the treatment of knowledge and time has also received a lot of attention in the form of variants of Alternating Temporal Epistemic Logic, e.g.,  \cite{penczek2003verifying,vanderHoek:2002:TMP:545056.545095}.)

Also adopting the philosophy of trading off expressive power with computational efficiency, the EL-O framework takes a syntactic restriction on the general problem of epistemic planning \cite{CooperHMMR16,CooperHMPV20,cooper2021lightweight}. Similar to our approach, EL-O is restricted to conjunctions of epistemic literals and the representation is compiled to a planning formalism. The key distinctions between our work and that of Cooper \emph{et al.} are: (1) they choose ``knowing whether'' as a primitive; (2) our style of compilation  (i.e., using ancillary effects) is unique; and, most crucially, (3) we capture a form of doxastic reasoning whereas they capture a restricted form of S5 epistemic knowledge. This final point is a fundamental distinguishing factor between the approaches -- but we note there are situations where either one would be the preferred setting.

As mentioned above, our technique was  inspired by recent,  state-of-the-art approaches to  planning  with partial observability \cite{BrafmanS12,BonetG14}. Of course, planning  with partial observability is a flavor of epistemic planning --  in that the uncertainty captures an implicit belief state -- 
albeit a limited one in which beliefs are not nested. Approaches such as \cite{BonetG14} consider the problem of how beliefs can be represented as classical states, by ``compilation.'' Thus, from an epistemic planning standpoint, only  individual knowledge of facts about the world (as opposed to belief) are handled:  the agent can ``know $p$ holds'' (i.e., $Kp$), ``know $p$ does not hold'' (i.e., $K\neg p$), or ``not know the value of $p$'' (i.e., $\neg K p \wedge \neg K \neg p$). (The multi-agent case is not considered.) 
The use of a knowledge modality was extended to be predicated on assumptions about the initial state, leading to effective techniques for conformant and contingent planning \cite{PalaciosG09,AlborePG09}.

\citet{kominisbeliefs} also take their inspiration from this lineage, and is the work that is most related to ours. They too share the  general motivation of bridging the rich fields of epistemic reasoning and automated planning by using classical planning over multi-agent epistemic states. However, the two approaches are fundamentally different and as a result each comes with its own strengths and shortcomings. The largest difference is our choice to focus on belief rather than knowledge -- for us, modeling the possibility of incorrect belief is essential. In contrast, \citet{kominisbeliefs} assume that all agents start with common initial knowledge, and further assume that all action effects are commonly known to all agents. (We can easily incorporate  this setting as a special case, but it is not necessary.) Conversely, they are able to handle arbitrary formulae, including disjunctive knowledge, while we are restricted to reasoning with RMLs. Moving forward, we hope to explore how we can combine ideas from both approaches.

The restriction on our specification language builds on a restricted fragment studied in \cite{Lakemeyer}. They introduce so-called \emph{Proper Epistemic Knowledge Bases} (PEKBs), where disjunctive knowledge is not permitted, enabling computationally tractable reasoning. We leverage that fragment in that the preconditions, goals, and states of our work can be viewed as PEKBs. In that spirit, approaches such as the 0-approximation semantics \cite{BaralS97} are alternate candidates for achieving tractability in reasoning. See, for example, \citet{son2017answer}. Our approach can be seen as having the ``Epistemic Closed World Assumption'' \cite{DBLP:conf/ijcai/WanYFLX15}. In essence, after compilation, the states contain explicit representation of everything that can be proved (restricted to our language).
\label{page:ecwa}

\section{Concluding Remarks}
\label{sec:conclusion}
We have presented a model of planning with nested belief, and the key contribution of this paper is to show that epistemic planning within this model can be done efficiently. 

We have demonstrated how a syntactically restricted class of problems involving planning with nested belief can be compiled into classical planning problems.  Despite the restricted form, we are able to model complex phenomena such as public or private communication, commonly observed action effects, and non-homogeneous agents (each with their own view of how the world changes).
Our focus on belief (as opposed to knowledge) provides a realistic framework for an agent to reason about a changing environment where knowledge cannot be presumed.

To solve this expressive class of problems, we appeal to existing techniques for dealing with ramifications, and compile the problem into a form that classical planning can handle. We show that our approach can solve a wide variety of epistemic planning scenarios, from toy problems in the literature to larger-scale problems that contain longer plans -- indeed, for problems that can be modeled in the language of other epistemic planners, we solve them more efficiently by orders of magnitude. Further, since we use classical planning as a black-box technology, as the field progresses, so does the strength of our approach. We have illustrated as such by comparing the performance of our approach using a modern classical planner versus one from only a few years ago.

In the future, we hope to expand the work in three key directions. First, we would like to explore other forms of ancillary conditional effects similar to the conditioned mutual awareness to give the designer greater flexibility during modelling (e.g., with concepts such as teamwork protocols or social realities). Second, we want to formalize the connection between general multi-agent epistemic planning and the syntactic restriction that we focus on encoding. We hope to provide an automated sound (but incomplete) approximation of an arbitrary \fullepplan problem into a \epplan problem. Finally, we would like to increase the expressiveness of our problems, in particular, by introducing a restricted form of disjunctive reasoning allowing modellers to express the concept of an agent $a$ \emph{knowing whether} agent $a$ knows $p$ is true, without agent $a$ knowing whether $p$ is true or not; for example, agent $a$ sees agent $b$ look at the clock, so $a$ knows that $b$ knows whether it is passed midnight, but agent $a$ does not know themselves. While more general disjunction is more expressive, it is more computationally challenging. `Knowing whether' is by far the most common type of disjunctive knowledge we have encountered in our applications.

  \subsubsection*{Acknowledgements}
\label{sec:acknowledgements}

\noindent This research is partially funded by Australian Research Council Discovery Grant DP130102825, \emph{Foundations of Human-Agent Collaboration: Situation-Relevant Information Sharing}, and by the Natural Sciences and Engineering Research Council of Canada (NSERC).
We would like to also thank Biqing Fang, Yongmei Liu, Son Tran, and Francesco Fabiano for extensive help in understanding their systems and the differences between them and RP-MEP. We also thank Maayan Shvo for his help in characterizing the space of existing epistemic planners.

  \bibliographystyle{elsarticle-harv}
  \bibliography{references}

  \newpage
  
  \appendix

\section{Glossary of Key Terms}
\label{app:terms}

\noindent
\begin{figure}[h!]
\small
\begin{tabularx}{\textwidth}{ l p{0.28\textwidth} l p{0.43\textwidth}}
  Acronym & Extended form & Sec. & Description \\
\hline 
RML & restricted modal literal & \ref{sec:background:pekbs} & literals of the form \newline $\phi ::= p \mid \belief{i} \phi \mid \neg \phi$ \\
PEKB & proper epistemic \newline knowledge base &  \ref{sec:background:pekbs} & a set of RMLs  \\
PINF & prime implicate \newline normal form & \ref{sec:background:pekbs} & a PEKB representation with polynomial entailment  \\
FOND & fully-observable  non-\newline deterministic planning & \ref{sec:classical-fond-planning} & classical planning, with non-deterministic operators \\
MEP & multi-agent epistemic \newline planning & \ref{sec:rpmep} & planning problem with bounded-depth RMLs as fluents \\
RP-MEP & restricted-perspective \newline multi-agent planning & \ref{sec:rpmep} & a MEP with a root agent and no negated belief (RMLs $\neg \belief{i} \phi$)\\
AK & always known fluents & \ref{sec:always-known} & there is common knowledge about the value of these fluents  \\
\hline
\end{tabularx}
\end{figure}

\section{Exemplary Domains}
\label{app:exemplary-domains}

Here we detail two domains in the file format used by RP-MEP -- the PDKB Domain Description Language (PDKBDDL). The language is a variant of the Planning Domain Definition Language (PDDL) \cite{pddlbook} which allows for the expressiveness of nested agent belief, conditioned mutual awareness, always known fluents, etc. The first domain demonstrates some of the features of the language using the grapevine domain as an example. The second demonstrates the drawback of RP-MEP not being able to reason with disjunctive beliefs.

\subsection{Grapevine}
The following snippets of PDKBDDL demonstrate the features of the specification language beyond PDDL, and corresponds to the Grapevine problem listed in Section \ref{sec:evaluation} where there are 4 agents, 2 goals, and depth 2 reasoning. Note that while we list four separate examples, they correspond to the same problem and are connected through the \verb|include| functionality.

\begin{labeledpdkbddl}[prob-4ag-2g-2d.pdkbddl]
{include:domain-4ag.pdkbddl}

(define (problem prob-4ag-2g-2d)

    ; PDKBDDL allows for including files in order to
    ;  compose common elements.
    {include:problem-setup-2d.pdkbddl}

    ; This indicates the restricted depth of nesting
    ;  that will be compiled
    (:depth 2)

    ; [ag]XYZ corresponds to agent ag believing XYZ
    
    ; <ag>XYZ corresponds to agent ag thinking XYZ is possible
    
    ; (!fluent) is used in lieu of (not (fluent)) to make
    ;  the task of parsing easier.
    
    (:goal
        [b][c](!secret a)
        [c](secret a)
    )
)
\end{labeledpdkbddl}

\begin{labeledpdkbddl}[domain-4ag.pdkbddl]
(define (domain grapevine)

    ; This specifies the finite number of agents
    (:agents a b c d)

    {include:domain.pdkbddl}
    
)
\end{labeledpdkbddl}

\begin{labeledpdkbddl}[domain.pdkbddl]
    (:types loc)
    (:constants )

    ; Predicates marked with {AK} are "Always Known".
    ;  The remaining predicates will be believed to
    ;  some nesting by the agents.
    (:predicates
            (secret ?agent)
        {AK}(at ?agent - agent ?l - loc)
        {AK}(connected ?l1 ?l2 - loc)
        {AK}(initialized)
    )

    (:action move
    
        ; The derive-condition specifies the condition for
        ;  mutual awareness. "always" translates to True,
        ;  while "never" translates to False.
        :derive-condition   always
        
        :parameters         (?a - agent ?l1 ?l2 - loc)
        
        :precondition       (and (at ?a ?l1)
                                 (connected ?l1 ?l2)
                                 (initialized))
                                 
        ; Note again that we use (!at ...) rather than
        ;  the common PDDL style of (not (at ...))
        :effect             (and (at ?a ?l2) (!at ?a ?l1))
    )


    (:action share
    
        ; This condition stipulates that agents are aware
        ;  of this action when they are "at" the location.
        ;  The parameter ?l is bound to the ?l listed in
        ;  the :parameters section, and $agent$ is a stand-in
        ;  for every agent in the domain.
        :derive-condition   (at $agent$ ?l)
        
        :parameters         (?a ?as - agent ?l - loc)
        
        ; Note that belief can be part of the precondition.
        :precondition       (and (at ?a ?l)
                                 (initialized)
                             [?a](secret ?as))
        
        ; Quantification will include all agents, including
        ;  the acting one.
        :effect         (and
                            (forall ?a2 - agent
                                (when
                                   (and      (at ?a2 ?l)
                                        <?a2>(secret ?as))
                                    [?a2](secret ?as)))
                        )
    )

    (:action fib
        :derive-condition   (at $agent$ ?l)
        :parameters         (?a ?as - agent ?l - loc)
        :precondition       (and     (at ?a ?l)
                                     (initialized)
                                 [?a](secret ?as))
        :effect         (and
                          (forall ?a2 - agent
                                (when
                                   (and      (at ?a2 ?l)
                                        <?a2>(!secret ?as))
                                   [?a2](!secret ?as)))
                        )
    )

    (:action initialize
        :derive-condition   never
        :precondition       (and)
        :effect             (and
                                (initialized)
                                (forall ?ag - agent
                                    [?ag](secret ?ag))
                            )
    )
\end{labeledpdkbddl}

\begin{labeledpdkbddl}[problem-setup-2d.pdkbddl]
    (:domain grapevine)

    (:objects l1 l2 l3 - loc)

    ; This allows us to project to individual agents, and is
    ;  not discussed in this paper.
    (:projection )
    
    ; The task of valid_generation is to create a plan. To
    ;  confirm a plan instead, valid_assessment can be used,
    ;  along with a list of actions in a :plan field.
    (:task valid_generation)

    ; The :init-type indicates the assumption of the root
    ;  agent. Here, it means that every RML not listed is
    ;  presumed to be possible.
    (:init-type complete)
        (:init

        ; Map
        (connected l1 l2)
        (connected l2 l1)
        (connected l2 l3)
        (connected l3 l2)

        ; Agents all in l1
        (forall ?ag - agent (at ?ag l1))

        ; Agents all believe others think secrets are possible
        (forall ?ag1 - agent
          (forall ?ag2 - agent
            (forall ?s - agent
                (and
                  [?ag1]<?ag2>(secret ?s)
                  [?ag1]<?ag2>(!secret ?s)
                ))))
    )
\end{labeledpdkbddl}

\subsection{Envelope}
The structure of this domain is that two agents observe a secret in an envelope, and they update their nested belief about the truth/falsehood of that secret. We consider the viewpoint of both the first and second agent to open the envelope in this setting. In PDKBDDL:

\begin{pdkbddl}
(define (domain envelope)

    (:agents alice bob)
    (:types )
    (:predicates (secret) )

    (:action check
        :derive-condition always
        :parameters       (?ag - agent)
        :precondition     (and)
        :effect           (and (when  (secret) [?ag](secret))
                               (when (!secret) [?ag](!secret)))
    )
)

(define (problem future-reasoning)
    (:domain envelope)
    (:projection )
    (:depth 2)
    (:task valid_assessment)

    (:init-type complete)
    (:init
        (forall ?ag1 - agent (and
            <?ag1>(secret)
            <?ag1>(!secret)
            (forall ?ag2 - agent (and
                [?ag2]<?ag1>(secret)
                [?ag2]<?ag1>(!secret)))))
        (secret)
    )

    (:goal (and [bob][alice](secret)))

    (:plan
        (check bob)
        (check alice)
    )
)
\end{pdkbddl}

Notice that the agents begin believing the secret (and its negation) is possible. Also, they believe that the other agents thinks it possible as well. The action to check is extremely compact and simple: for the agent that checks, they'll learn the true value of the secret.

With the mutual awareness aspect of RP-MEP, we have a further ramification that if the agent believes the secret (resp. its negation) when the other agents checks, then they'll come to believe the other agent believes (resp. doesn't believe) it as well. The syntax used above is to verify that the plan [(check bob), (check alice)] achieves the goal of [bob][alice](secret).

This is readily handled by RP-MEP. After the first action, bob believes the secret, and through the ancillary effects bob comes to believe alice does as well after the second action. However, the reversed plan no longer works:

\begin{pdkbddl}
  ...
    (:plan
        (check alice)
        (check bob)
    )
  ...
\end{pdkbddl}

Here, bob would need to retain the information that alice believes either the secret or its negation, and then reconcile that disjunctive fact with their own discovery of the true value. We also see this phenomenon in the Hattari domain.

\section{Encoding Ancillary Effects}
\label{app:ancillary-domains}

Here we demonstrate the encoding of the core concepts of RP-MEP -- ancillary effects -- in both our modelling language (PDKBDDL), as well as that of MEPK (EPDDL). Most of the ancillary effects that we handle as part of the preprocessing are natively handled by the MEPK planner, and these examples serve to illustrate the commonalities.
%

\subsection{Negation Removal}
The first example is a simple domain to demonstrate the impact of negation removal, corresponding to Eqn (\ref{eq:remove-negation}). We list the entire domain and problem for both languages, but subsequent examples will just show the relevant action descriptions for the particular phenomenon.

\begin{pdkbddl}
(define (domain negation-removal)
    (:agents a)
    (:types )
    (:constants )
    (:predicates (p) (q))

    (:action apply
        :derive-condition   always
        :precondition       (and )
        :effect             (and [a](p))
    )

    (:action check
        :derive-condition   always
        :precondition       (and (not <a>(!p)))
        :effect             (q)
    )
)

(define (problem prob)
    (:domain negation-removal)
    (:projection )
    (:depth 2)
    (:task valid_generation)
    (:init-type complete)
    (:init
        <a>(p)
        <a>(!p)
    )
    (:goal (q))
)
\end{pdkbddl}

Note that the general structure of these examples are to make the goal fluent (q) true, which can only occur in the action sequence [apply,check]. It is the `check' action that checks (through its precondition) that the effect of `apply' is correctly handled. Here is the equivalent problem for EPDDL used by MEPK:

\begin{epddl}
(define (domain negation-removal)
    (:objects)
    (:agents a )
    (:predicates  (p) (q) )
   
    (:action apply
        :category (ontic)
        :parameters ()
        :precondition (True)
        :effect (<{(True)} {(K_a (p))}>))
    (:action check
        :category (ontic)
        :parameters ()
        :precondition (not (DK_a (not (p))))
        :effect             (<{(True)} {(q)}>))
    
    (:init (and (DK_a (p)) (DK_a (not (p)))))
    (:constraint (True))
    (:goal (q))
)
\end{epddl}

The correspondence between the languages is fairly direct. For every `[agent]' in PDKBDDL, we have `K\_agent' in EPDDL, etc. Both planners successfully handle this problem, as it only requires removing the negated information of an effect from the knowledge base.

\subsection{Closure}
Next, we consider the application of logical closure to effects (cf. Eqn (\ref{eq:closure})).

\begin{pdkbddl}
  ...
    (:action apply
        :derive-condition   always
        :precondition       (and )
        :effect             (and [a](p))
    )

    (:action check
        :derive-condition   always
        :precondition       (and <a>(p))
        :effect             (q)
    )
  ...
\end{pdkbddl}

\begin{epddl}
  ...
    (:action apply
        :category (ontic)
        :parameters ()
        :precondition (True)
        :effect (<{(True)} {(K_a (p))}>)
    )

    (:action check
        :category (ontic)
        :parameters ()
        :precondition (DK_a (p))
        :effect       (<{(True)} {(q)}>)
    )
  ...
\end{epddl}

Again, we find that both RP-MEP and MEPK readily handle this situation.

\subsection{Inverted Closure}
Corresponding to Eqn (\ref{eq:uncertainty-closure}), the inverted closure ensures that the knowledge base doesn't contain information that would entail something we are removing. Once again, both RP-MEP and MEPK handle this situation equally well.

\begin{pdkbddl}
  ...
    (:action apply
        :derive-condition   always
        :precondition       (and )
        :effect             (and (not <a>(p)))
    )

    (:action check
        :derive-condition   always
        :precondition       (and (not [a](p)))
        :effect             (q)
    )
  ...
    (:init [a](p) )
  ...
\end{pdkbddl}

\begin{epddl}
  ...
    (:action apply
        :category (ontic)
        :parameters ()
        :precondition (True)
        :effect (<{(True)} {(not (DK_a (p)))}>)
    )

    (:action check
        :category (ontic)
        :parameters ()
        :precondition (not (K_a (p)))
        :effect             (<{(True)} {(q)}>)
    )
  ...
    (:init (K_a (p)))
  ...
\end{epddl}

\subsection{Uncertain Firing}
The final ancillary effect, corresponding to Eqn (\ref{eq:uncertainty-ce}), highlights where the two approaches diverge. ``Uncertain firing'', as identified in the text, is a known phenomenon for belief update in action theories with partial observability. The distinguishing factor is that this inference \textit{involves reasoning about how an action affects the world}. The previous three examples simply refer to ramifications of maintaining a consistent \kd knowledge base.

From the PDKBDDL, notice that the agent begins thinking (q) must be false, but then comes to think it may be possible.
\vspace{2em}

\begin{pdkbddl}
  ...
    (:action apply
        :derive-condition   always
        :precondition       (and )
        :effect             (and (when (p) (q)))
    )

    (:action check
        :derive-condition   always
        :precondition       (and <a>(q))
        :effect             (r)
    )
  ...
    (:init [a](!q) )
    (:goal (r))
  ...
\end{pdkbddl}

The corresponding EPDDL, with an added effect to ensure that the agent knows about the conditional effect (i.e., ``K\_a (p) $\rightarrow$ K\_a (q)'') is as follows:

\begin{epddl}
  ...
    (:action apply
        :category (ontic)
        :parameters ()
        :precondition (True)
        :effect (<{(p)} {(q)}>
                 <{(K_a (p))} {(K_a (q))}>)
    )

    (:action check
        :category (ontic)
        :parameters ()
        :precondition (DK_a (q))
        :effect       (<{(True)} {(r)}>)
    )
  ...
    (:init (K_a (not (q))))
    (:goal (r))
  ...
\end{epddl}

MEPK does not handle this case. The reason being is that any ramification that involves \textit{knowledge about the impact an action has on the world is not captured by the planner natively}. In order to cover the case of uncertain firing, the ramification must be written manually. This would correspond to the domain author specifying an effect of the form:

\begin{epddl}
    <{DK_a (p)} {DK_a (q)}>
\end{epddl}

While this is a minor modification to make the domain work correctly with MEPK, note that it is a ramification of sorts that needs to be covered automatically. Otherwise, all ramifications become essentially those that must be written by hand. See Section \ref{sec:mepk-comparison} for further discussion.

While we do not detail an example of conditioned mutual awareness, the impact is the same. MEPK does not have a native treatment of such phenomenon (i.e., subsets of agents that are mutually aware of the impact an action has), and thus must be modelled by hand.

\section{PEKBs for Efficient Planning}
\label{app:pekbs}

Here, we outline a theory of Proper Epistemic Knowledge Bases (PEKB) that are suitable for epistemic planning in a KD$_n$/KD45$_n$ context.

\subsection{PEKBs}


In this work, we expand the theory of PEKBs for the logic of \kd and \kdff, suitable for representing epistemic planning problems on top of STRIPS-based planners. The choice of PEKBs is suitable for two reasons:

\begin{enumerate}
    \item The syntactic restrictions imposed by PEKBs that prevent disjunction and infinite nesting are consistent with the syntactic restrictions employed by STRIPS-based planners, in which arbitrary disjunction is not permitted, and the set of literals (fluents) in a planning problem are finite. In this sense, using PEKBs increases the expressiveness of the STRIPS language to include epistemic formulae.
    \item PEKBs come with nice logical and computational properties, as we show here. First, a \emph{consistent} PEKB --- that is, one with no contradictory statements --- is \emph{logically separable}, which means the literals in the knowledge base do not interact to produce new formulae. Further, a consistent PEKB can be queried in polynomial time without a pre-compilation step such as the one used by \citeasnoun{Lakemeyer}, and can be queried in constant time with an exponential pre-compilation step. Finally, a consistent PEKB can be updated with new literals and remain consistent using a polynomial-time algorithm.
\end{enumerate}

Thus, given these two above properties, PEKBs make a suitable representation for extending classical planning over belief bases.
In the remainder of this appendix, we prove these properties of PEKBs and analyse their complexity.

\subsection{Logical Separability in PEKBs}

The property of logical separability of formulae in PEKBs is a key property in the complexity analysis of PEKBs.

\begin{definition}{Logical Separability \cite{Lakemeyer}}
\label{defn:logical-separability}
The set of RMLs $P$ is \emph{logically separable} if and only if for every consistent set of RMLs $P'$ the following holds:
\begin{align*}
\textrm{if} \mm{3} P \cup P' \models \bot \mm{3} \textrm{then} \mm{3} \exists \literala \in P, \textrm{ s.t. } P' \cup \set{\literala} \models \bot
\end{align*}
\end{definition}

Intuitively, a set of formulae is logically separable if we cannot infer anything by combining two or more formulae from the set. E.g., $\set{\belief{i}p, \belief{i}(p \supset q)}$ is not logically separable, because we can infer $\belief{i}q$ from the combination of the two formulae in the set. The set $\set{\belief{i}p, \belief{i}(p \supset q)} \cup \set{\possible{i}\neg q}$ is inconsistent, but $\possible{i}\neg q$ is consistent with both other formulae individually. Logical separability plays an important role later, as examples such as the one above are forbidden. The core issue is the use of disjunction (manifest in the example as an implication) which opens the door to case-based reasoning and far more complex issues when it comes computing all possible ramifications.


To simplify the notation, we will denote a PEKB as a conjunction

\begin{align*}
       &\Gamma \\
\land\mm{2} &\possible{i}\psi_1^i \land \ldots \land \possible{i}\psi_m^i \land \belief{i}\chi_1^i \land \ldots \land \belief{i}\chi_n^i \\
\land\mm{2} &\possible{j}\psi_1^{j} \land \ldots \land \possible{j}\psi_m^{j} \land \belief{j}\chi_1^{j} \land \ldots \land \belief{j}\chi_n^{j} \\
\land\mm{2} &\cdots
\end{align*}

Here, $\Gamma$ is a conjunctive propositional formula, and we have grouped the $\belief{}$ and $\possible{}$ operators for each particular agent so that each $\psi$ and $\chi$ symbols can be used to unambiguously identify the outermost operator.

\begin{theorem}
\label{thm:unsatisfiable-pekb}
Given a \kb $P$ in \kd assumed to be in the form above, we have that $P \models \bot$
$\mathit{iff}$ at least one of the following holds:
\begin{enumerate}[(a)]\itemsep=0pt
 \item $\Gamma \models \bot$
 \item $\psi_k^i \land \chi_1^i \land \ldots \land \chi_n^i \models \bot$ (for some agent $i$ and index $k$)
 \item $\chi_1^i \land \ldots \land \chi_n^i \models \bot$ (for some agent $i$).
\end{enumerate}
\end{theorem}
\begin{proof}
A proof for the logic $K_n$ (for which only parts (a) and (b) above are required) is presented by \citeasnoun{bienvenu09} (see Theorem~1, part (3)) for the single agent case. It is straightforward to see that with the addition of the axiom $D$, that part (c) must be added as a consistent \kb cannot contain the formula $\belief{i}\bot$.
Finally, we note that \citeauthor{bienvenu09}'s \cite{bienvenu09} proof extends to the multi-agent case \kd because two or more agents can believe contradictory propositions in \kd --- it is only an agent's internal beliefs that must be consistent.
\end{proof}

Following directly from \kd, we can use the following lemma (cf. \cite{hughes1996new,Chellas80}):

\begin{lemma}
\label{lemma:nec}
$\varphi \models \psi ~\mathit{implies}~ 
 \possible{i}\varphi \models \possible{i}\psi ~\mathit{and}~
 \belief{i}\varphi \models \belief{i}\psi$
\end{lemma}

\begin{theorem}
\label{thm:logically-separable-pekb}
Given a consistent \kb $P$ and RML $\psi$ in \kd, if $P \models \psi$ then $\exists \literala \in P$, s.t.\ $\literala \models \psi$
\end{theorem}
\begin{proof}
We prove this inductively on the structure of $\psi$. Assume $P=(\Gamma \land \possible{i}\psi_1 \land \ldots \land \possible{i}\psi_m \land \belief{i}\chi_1 \land \ldots \land \belief{i}\chi_n)$. 

The case of $\psi \equiv \gamma$, with $\gamma$ being some conjunct in $\Gamma$, is straightforward because \kbs contain no disjunction. For the $\psi \equiv \belief{i}\psi'$ case, $P \models \belief{i}\psi'$ iff $P \land \possible{i}\neg\psi' \models \bot$. From Theorem~\ref{thm:unsatisfiable-pekb} and the assumption that the \kb is consistent, it follows that $\schi \models \psi'$. By induction, there must be some $\chi_k \in \set{\chi_1, \ldots, \chi_n}$ such that $\chi_k \models \psi'$. From Lemma~\ref{lemma:nec}, we know that $\belief{i}\chi_k \models \belief{i}\psi'$ and that $\belief{i}\chi_k \in P$, so this case holds.

The case of $\psi \equiv \possible{i}\psi'$ is similar. If $P \models \possible{i}\psi'$, then from  Theorem~\ref{thm:unsatisfiable-pekb}, it follows that either for some $\psi_j \in \set{\psi_1, \ldots, \psi_m}$, we have that $\psi_j \land \schi \models \psi'$, or that $\schi \models \psi'$. By induction, it must be that $\psi_j \models \psi'$ or $\chi_k \models \psi'$ for some $\chi_k \in \set{\chi_1, \ldots, \chi_n}$. From Lemma~\ref{lemma:nec} and axiom D, we know that either $\possible{i}\psi_j \models \possible{i}\psi'$ (where $\possible{i}\psi_j \in P$), or $\belief{i}\chi_k \models \possible{i}\psi'$ (where $\belief{i}\chi_k \in P$); so this case holds. From the three cases, the theorem holds.
\end{proof}

From the above theorem, it is clear to see that if a \kb is inconsistent, this inconsistency can be detected by checking all pairwise RMLs. Theorem \ref{thm:logically-separable-pekb} provides us with a very direct and tractable procedure for inference with PEKBs. The following corollary follows directly from Definition~\ref{defn:logical-separability} and Theorem~\ref{thm:logically-separable-pekb}.

\begin{corollary}
\label{cor:consistent-pekb-is-separable}
A consistent \kb in \kd is logically separable.
\end{corollary}

Corollary~\ref{cor:consistent-pekb-is-separable} is an important property in the context of the computational complexity of \epplan: given a \kb $P$ in \kd with $n$ elements, to update it with a new RML $\literala$, we only need to check $\literala$ against each element in $P$ to check if it is inconsistent with the belief base, which is in $O(n\cdot d)$ (with $d$ being the depth of the RML). Thus, detecting inconsistency (and performing belief update), requires only a pairwise check of all elements in the \kb, which has time complexity of just $O((n\cdot d)^2)$.

\subsection{Entailment in Consistent PEKBs}
\label{sec:pekb:entailment}

For definition purposes, we consider the space of all \kb~'s as a partially-ordered set (poset) $(R, \models)$, in which $R$ is the set of all RMLs. We use $P$ and $Q$ to refer to subsets of $R$ (that is, $P$ and $Q$ are PEKBs), and we use $\negkb{P}$ to denote the \kb that contains the negation of every RML in $P$; that is
$\negkb{P} = \set{\neg \literala \st \literala \in P}$. Figure~\ref{fig:pekb-lattice} shows a Hasse diagram of a poset. The poset is bounded, with the top element $\possible{}\possible{}\possible{}p$ and the bottom element $\bbb$. We only focus on posets across RMLs \textit{with the exact same prefix of agent modalites}. That is, sequences of either $\belief{}$ or $\possible{}$ operators that use the some ordering of agents. The appropriateness of this simplification will become clear in Definition \ref{defn:pekb-entailment} and Theorem \ref{thm:pekb-entailment}.

\begin{figure}[t!]
\centering
\hspace{-8mm}
\tikzstyle{bag} = [text width=2.5em, text centered]
\tikzstyle{child} = [text width=3em, text centered, fill = gray!40, shape = rectangle, rounded corners]
\begin{tikzpicture}
   \begin{scope}[node distance=2cm,on grid,>=stealth']
   \node [bag] (bbb)			  {$\bbb$};
   \node [bag] (bpb)	[above=of bbb]    {$\bpb$}  edge [-] (bbb);
   \node [child] (pbb)	[left=of bpb]     {$\pbb$}  edge [-] (bbb);
   \node [bag] (bbp)	[right=of bpb]    {$\bbp$}  edge [-] (bbb);
   \node [bag] (pbp)	[above=of bpb]    {$\pbp$}  edge [-] (pbb) edge [-] (bbp);
   \node [bag] (ppb)	[left=of pbp]     {$\ppb$}  edge [-] (pbb) edge [-] (bpb);
   \node [bag] (bpp)	[right=of pbp]    {$\bpp$}  edge [-] (bpb) edge [-] (bbp);
   \node [bag] (ppp)	[above=of pbp]    {$\ppp$}  edge [-] (ppb) edge [-] (pbp) edge [-] (bpp);
   \node[ellipse,rotate=-30,draw=black,fill=gray!40, opacity=0.3, dash pattern=on 1pt, fit=(bbb) (bbp) (bpb) (bpp), inner sep=-4.3mm](FIt2) {};
   \end{scope}
\end{tikzpicture}
\caption{A representative Hasse diagram with bottom element $\bbb$. The shaded set represents $\dclose{\bpp}$, which would be removed if $\bpp$ was erased from a \kb containing $\bbb$. The shaded formula $\pbb$ should be the RML that remains, because it is the maximal RML in $\close{\bbb}$ that is not in with $\dclose{\bpp}$.\vspace{-1em}}
\label{fig:pekb-lattice}
\end{figure}

\begin{definition}{Upwards and downwards closure}
Given an RML $\literala$, we define  the \emph{upward closure} $\close{\literala}$ of $\literala$ as the upward closure with respect to its poset, defined as:
\[
  \close{\literala} = \set{\literalb \mid \literala \models \literalb }
\]
The \emph{downward closure} of $\lit$, denoted $\dclose{\literala}$, is just $\negkb{\close{\neg\literala}}$; that is  $\dclose{\literala} = \set{\literalb \mid \literalb \models \literala}$.

 The upward closure of a \kb $P$ is $\close{P} = \bigcup_{\literala \in P}\close{\literala}$. The downward closure of a \kb $P$ is defined as $\dclose{P} = \bigcup_{\literala \in P} \dclose{\literala}$, or alternatively $\dclose{P} = \negkb{\close{\negkb{P}}}$.
\label{defn:up-down-closure}
\end{definition}

Intuitively, the upward closure of a PEKB $\close{P}$ is all RMLs that are entailed by $P$; and the downwards closure $\dclose{P}$ is all RMLs that entail an RML in $P$. Thus, $\close{\belief{i}p} = \{\belief{i}p, \possible{i}p\}$; and $\dclose{\possible{i}p} = \{\belief{i}p, \possible{i}p\}$.

We introduce the new notation for closure here to unify the notation of upward/downward closure that the main text did not require ($\close{}$ is analogous to $\closure{}$, and $\dclose{}$ is newly introduced). The upward or downward closure of any PEKB is finite. The logical separability of PEKBs means that combining two RMLs cannot produce any new RMLs that we cannot get from inferring from just one. For each RML, the number of RMLs that it entails is finite: we can only repeatedly apply the axiom D, thus traversing up the poset lattice, until we reach the top element.

\begin{definition}{Algorithm for PEKB Entailment in $KD_n$}


%
Given a consistent PEKB $P$ and a query $\varphi$ that is a conjunction of RMLs, we define \kd entailment $P \pekbentails \varphi$ as:

\begin{center}
\begin{tabular}{llll}
 & $P \pekbentails p$ & iff &  $p \in P$\\
 & $P \pekbentails \varphi \land \psi$ & iff & $P \pekbentails \varphi$ and $P \pekbentails \psi$\\
 & $P \pekbentails \neg \varphi$ & iff & $P \not\pekbentails \varphi$\\
 & $P \pekbentails \belief{i}\varphi$ & iff & for some $\belief{i}\psi \in P$, $\{\psi\} \pekbentails \varphi$\\
 & $P \pekbentails \possible{i}\varphi$ & iff & for some $\belief{i}\psi \in P$, $\{\psi\} \pekbentails \varphi$ or\\
 &                                &     & for some $\possible{i}\psi \in P$, $\{\psi\} \pekbentails \varphi$
 \end{tabular}
\end{center}
\label{defn:pekb-entailment}
\end{definition}

\begin{theorem}
The entailment algorithm defined by $\pekbentails$ is sound and complete in \kd: i.e., $P \pekbentails \varphi$ iff $P \vDash \varphi$, where $\vDash$ is the standard entailment defined in Section~\ref{sec:preliminaries}.
\label{thm:pekb-entailment}
\end{theorem}

\begin{proof}
The cases for the propositional primitive, conjunction and negation are straightforward from their definition.
Given that we assume the query is a conjunction of RMLs, the conjunction in Definition \ref{defn:pekb-entailment} is the only inductive argument required -- this follows the standard notion of entailment for a conjunctive query. Therefore, we must only consider $P \pekbentails \varphi$ where $\varphi$ is an RML.

In this case, we can appeal to Theorem \ref{thm:correct-closure} (capturing the ways one RML entails another) and Theorem \ref{thm:logically-separable-pekb} (capturing logical separability) to ensure that $P \pekbentails \varphi$ if and only if $P \entails{} \varphi$, where $\varphi$ is an RML.


\end{proof}

\begin{theorem}
Entailment for consistent PEKBs in \kd has worst-case complexity of $O(|P| \cdot |\varphi| \cdot d)$, in which $|P|$ is the PEKB size, $|\varphi|$ is the number of conjuncts in the query, and $d$ is the longest RML in the PEKB or query.
\end{theorem}
\begin{proof}
Given the logical separability, we simply need to compare each element in the PEKB with each element of the conjunction in the query, which is $|P| \cdot |\varphi|$. For each query, the worst case is to compare along the length of the RML using the last two rules until a propositional literal is encountered, thus the worst case complexity is $O(|P| \cdot |\varphi| \cdot d)$.
\end{proof}

\begin{definition}{Entailment as closure}
\label{defn:entailment-as-closure}
Given a consistent PEKB $P$, and a query $\varphi$ in \kd that is a conjunction of RMLs, we define entailment as: $P \pekbentails \varphi$ iff for all $\psi \in \varphi$, $\psi \in \close{P}$. In other words, if we compute the closure of $P$, we need only check containment in the closure for every RML in the conjunction $\varphi$.
\end{definition}

The soundness and completeness of this holds trivially from the definition of closure. 

\begin{theorem}
Entailment as (upward) closure for consistent PEKBs has worst-case complexity of $O(|P| \cdot |\varphi| \cdot 2^d)$, in which $|P|$ is the size of the PEKB, $|\varphi|$ is the number of conjuncts in the query, and $d$ is the longest RML in the PEKB or query. 
However, if each RML is indexed using, for example, hashing when it is computed as part of a the closure, complexity is $O(|\varphi|)$ for any query once the closure is computed.
\end{theorem}
\begin{proof}
The size of $\close{P}$ is $|P| \cdot 2^d$, because for each RML of length $d$, there are $2^d$ RMLs that follow from it by applying axioms K and D. Thus, the closure is computed in $O(|P| \cdot 2^d)$ time, and then $\varphi$ queries must be run against each RML in $\close{P}$, which has a worst case of $O(|P| \cdot |\varphi| \cdot 2^d)$.
If each RML in $\close{P}$ is indexed using hashing and the depth is bounded, the lookup becomes constant, so the complexity is just looking up $|\varphi|$ number of queries. 
\end{proof}

Using indexing still requires a compilation of $\close{P}$, which has a worst-case complexity of $O(|P| \cdot 2^d)$. However, if there are multiple queries to be run against the knowledge base, using compilation and hashing as a way of ammortizing the cost is valuable. In Section~\ref{sec:encoding}, we showed how this compilation is encoded within a planning model, so the lookup provides significant value.

\subsection{Belief Erasure and Update in PEKBs}
\label{sec:pekbs:update}

Here, we outline a polynomial-time algorithm for belief update --- a key property required for PEKBs to be useful in planning.

Our definition of belief update in \kbs uses the `forget-then-conjoin' approach of first removing any beliefs that conflict with the update, and then adding the update. This `forgetting' process, which \citeasnoun{KM91} term \emph{belief erasure}, is not simply the act of subtracting the negation of the RMLs in the update, because we must also remove any RML that implies the negated update. For example, given the \kb $\set{\possible{i}p, \belief{i}p}$, removing $\possible{i}p$ should also remove $\belief{i}p$; otherwise, the belief base would still entail $\possible{i}p$. Further, a belief erasure operator should follow the principle of \emph{minimal change}: when removing belief from an existing belief base, we should remove only what we must so that the belief base no longer entails the removed belief.

\begin{definition}{Prime PEKBs}
A \kb $P$ is \emph{prime} if and only if all elements in $P$ are prime implicates of $P$ (maximal elements); that is, for all $\literala, \literalb \in P$, if $\literala \models \literalb$ then $\literalb \models \literala$. The set of maximal elements of a \kb is denoted $\max(P)$.
\end{definition}

Note that for any \kb $P$ (prime or otherwise) and an RML $\literala$, we have that $P \models \literala$ iff $\literala \in \close{P}$.

\begin{definition}{Belief erasure in \kbs}
\label{defn:belief-erasure}
Given \kbs $P$ and $Q$, we define $P \erase Q$ as the \emph{belief erasure} of  $Q$ from $Q$ as follows:
\begin{align*}
P \erase Q =  \max(\close{P} \setminus \dclose{Q})
\end{align*}
That is, take the upward closure of $P$ and remove the downward closure of $Q$, removing any non-prime RMLs. This removes $Q$ and anything that implies it, leaving those things that are in the upward closure of $P$ that do not entail $Q$.
\end{definition}

\begin{theorem}
The complexity of $P \erase Q$, where $P$ and $Q$ are both prime, is $O(|P| \cdot |Q| \cdot d)$, in which $d$ is the depth of the longest RML in $P$ or $Q$. If instead we calculate $P \erase Q$ by calculating  $\close{P}$ and $\dclose{Q}$, and the subtracting their difference, the complexity is $O(2^{|P|} \cdot 2^{|Q|} \cdot d)$.
\end{theorem}
\begin{proof}
Given the logical separability of PEKBs, we need to only calculate the erasure of each pair in $P \times Q$. This erasure can be done linearly in the depth of the RMLs. For each modal operator index that is $\belief{i}$ in both $\literala$ and $\literalb$, create a new RML that is equivalent to $\literala$ but with $\possible{i}$ at that index, and then take the top elements of this set. This can be calculated by traversing up the poset until we reach the top elements (see Figure~\ref{fig:pekb-lattice} for an example), which has a maximum depth of $d$.

If we instead calculate the upward and downward closure of $P$ and $Q$ respectively, we need to just iterate through the $2^{|P|} \times 2^{|Q|}$ pairs and remove any from $\close{P}$ that occurs in $\dclose{Q}$. Checking each pair has linear complexity in $d$, although as with entailment, this operation can be done in constant time with indexing.
\end{proof}

Given a belief erasure operator, belief update for a \kb is straightforward: update is forget (erase) then conjoin, eliminating any RML that is not prime.

\begin{definition}{Belief Update in PEKBs}
\label{defn:belief-update}
Given \kbs $P$ and $Q$, we define belief update of $P$ with $Q$, denoted as $P \update Q$, as follows:
\begin{align*}
P \update Q = \max((P \erase \negkb{Q}) \cup Q)
\end{align*}
That is, remove anything that conflicts with $Q$, then add the elements in $Q$, and take the maximal elements from the result.
\end{definition}

The complexity of this is just the complexity of belief erasure, with the added overhead of adding the new elements in $Q$ into the PEKBs.

This update operator observes the property of relevant minimal change \cite{perrussel2012relevant}. Because PEKBs are logically separable, it is clear to see that any RMLs that are not related to the new RMLs remain in the knowledge base.

Thus, we have defined belief erasure and belief update for \kbs.

\subsection{Analysis of Belief Update and Erasure Operators}

In this section, we extend the results by  \citeasnoun{miller-ijcai16} that analyse these operators with respect to the classic belief update postulates from \citeasnoun{KM91}. Specifically, we include proofs that did not appear in the published version. These results show that the belief update and erasure operators are satisfiable and correct.
 
\subsubsection{KM Postulates for Belief Update}

 \citeasnoun{KM91} propose a set of postulates for belief update called the Katsuno-Mendelzon (KM) postulates. These postulates, which echo the AGM postulates for belief revision \cite{AlchourronGM85}, specify eight properties that a belief update operator should have to be an appealing update mechanism (phrased using our notation):

\begin{description}
    \item[U1] $P \update Q \models Q$
    \item[U2] If $P \models Q$ then $P \update Q \equiv P$
    \item[U3] If $P$ and $Q$ are satisfiable, then $P \update Q$ is satisfiable
    \item[U4] If $P \equiv P'$ and $Q \equiv Q'$ then $P \update Q \equiv P' \update Q'$
    \item[U5] $(P \update Q) \conjoin R \models P \update (Q \conjoin R)$
    \item[U6] If $P \update Q \models R$ and $P \update R \models Q$ then $P \update Q \equiv P \update R$
    \item[U7] If $P$ is complete then $(P \update Q) \conjoin (P \update R) \models P \update (Q \vee R)$
    \item[U8] $(P \vee Q) \update R \equiv (P \update R) \vee (Q \update R)$
\end{description}

Because \kbs do not permit disjunction, U7 and U8 are not relevant for our belief update operator.

 Despite their widespread use, it is not commonly accepted that all postulates are desirable for all belief update operators. \citeasnoun{herzig1999propositional} argue that only postulates U1, U3, U8, and (possibly) U4 should be satisfied by all update operators. 

\begin{theorem}
KM postulates U1, and U3-U6 hold for \kb belief update operator $\update$. U2 holds if $P$ is satisfiable.
\end{theorem}

\begin{proof}
The proofs for U1 and U4 are straightforward.

U2: From the definitions of $\update$ and $\erase$,  $P \update Q$ is equivalent to $\max(\close{P} \setminus \dclose{\negkb{Q}}) \conjoin Q$. If $P \models Q$ and $P$ is satisfiable, we know that $\close{P} \cap \dclose{\negkb{Q}} = \emptyset$, and therefore $\close{P} \setminus \dclose{\negkb{Q}} = \close{P}$. Further, if $P \models Q$, we know that $\close{Q} \subseteq \close{P}$, and therefore, $\close{P} \cup \close{Q} = \close{P}$, and therefore $\max(\close{P} \cup \close{Q}) = \max(\close{P})$, meaning that postulate U2 holds.

U3: Assume that the postulate does not hold, so $P \not\models \bot$ and $Q \not\models \bot$, but $P \update Q \models \bot$. If this is the case, then there is some $\phi$ such that $\close{P} \setminus \dclose{\negkb{Q}} \models \phi$ and $\close{Q} \models \neg\phi$. However, from the definition of downwards closure, we know that $\dclose{\negkb{Q}} = \negkb{\close{Q}}$, meaning that $\close{P} \setminus \negkb{\close{Q}} \models \phi$. If $\close{Q} \models \neg\phi $ then it must be that $\negkb{\close{Q}} \models  \phi$, and as a result, $\close{P} \setminus \negkb{\close{Q}} \not\models \phi$, violating our assumption. Therefore, postulate U3 holds.

U5: Assume that the postulate does not hold. This implies that there is some $\phi$ such that $\close{P}\setminus\dclose{\negkb{Q \conjoin R}} \conjoin (Q \conjoin R) \models \phi$ and $\close{P}\setminus\dclose{\negkb{Q}} \conjoin (Q \conjoin R) \not\models \phi$. Because $Q \conjoin R$ occurs in the latter, $\phi$ cannot be entailed by $Q \conjoin R$. This implies that: (a) $\close{P}\setminus\dclose{\negkb{Q \conjoin R}} \models \phi$;  but (b) $\close{P}\setminus\dclose{\negkb{Q}} \not\models \phi$. From (a), we know that that $\close{P} \models \phi$, and so for (b)  to hold, it must be that $\dclose{\negkb{Q}} \models \phi$. However, $\dclose{\negkb{Q \conjoin R}} \models \dclose{\negkb{Q}}$, meaning that  $\close{P}\setminus\dclose{\negkb{Q \conjoin R}} \not\models \phi$, which contradicts (a). Therefore, postulate U5 holds.

U6: Assume that the postulate does not hold, then this means that if: (a) $\close{P}\setminus \dclose{\negkb{Q}} \conjoin Q \models R$ and (b) $\close{P} \setminus \dclose{\negkb{R}} \conjoin R \models Q$, then there exists some $\phi$ such that $\close{P} \setminus \dclose{\negkb{Q}} \conjoin Q \models \phi$ but $\close{P} \setminus \dclose{\negkb{R}}  \conjoin R \not\models \phi$ (or vice-versa, but the cases are symmetric). It cannot be that case that $Q \models\phi$, otherwise from (b) it would follow that $\close{P} \setminus \dclose{\negkb{R}}  \conjoin R \models \phi$, violating our assumption. Therefore, it must be that  $P \models \phi$ and $\negkb{Q} \not\models \phi$, and $(\close{P} \setminus \dclose{\negkb{R}}) \conjoin R \not\models \phi$. Because $P \models \phi$, it must be that $\dclose{\negkb{R}} \models \phi$, and therefore ${R} \models \neg\phi$. But from (a), this implies that $\close{P}\setminus \dclose{\negkb{Q}} \conjoin Q \models \neg\phi$, contradicting our assumptions. Therefore postulate U6 holds.
\end{proof}

\citeasnoun{KM91} present the so-called \emph{representation theorem}, which shows the completeness of the belief update operator. It is clear that the pre-order on interpretations defined by \citeauthor{KM91} can simply be defined over the poset corresponding to the elements in the \kb. Due to the logical separability of \kbs , Definition~\ref{defn:belief-update-and-erasure} amounts to an equivalent notion of \citeauthor{KM91}'s representation theorem.

\subsubsection{KM Postulates for Belief Erasure}

 \citeasnoun{KM91} also propose a set of postulates for belief erasure based on the principle of minimal change: when removing belief from an existing belief base, we should remove only what we must so that the belief base no longer entails the removed belief. These postulates phrased using our notation are:

\begin{description}

\item[E1] $P \models P \erase Q$ 
\item[E2] If $P \models \negkb{Q}$ then $P \erase Q \equiv P$
\item[E3] If $P$ is satisfiable then $P \erase Q \not\models Q$
\item[E4] If $P \equiv P'$ and $Q \equiv Q'$ then $P \erase Q \equiv P' \erase Q'$
\item[E5] $(P \erase Q) \conjoin Q \models P$
\item[E8] $(P \vee Q) \erase R \equiv (P \erase R) \vee (Q \erase R)$

\end{description}

E8 does not make sense because \kbs cannot contain disjunctive formulae.

\citeasnoun{KM91} define an identity, a mirror of the identity Harper introduced that expresses belief contraction in terms of set operations and belief contraction \cite{harper1976rational}. \citeauthor{KM91}'s identity can be expressed as:
\begin{align}
\label{eq:km-identity}
P \erase Q &\equiv P \disjoin (P \update \negkb{Q})
\end{align}

\noindent in which  $P \disjoin Q = \max(\close{P} \cap \close{Q})$.
Intuitively, this identity stipulates that erasing $Q$ should be the same as restricting the belief base to what would hold if the negation of $Q$ was added.

\citeasnoun{KM91} show that if the identity in Equation~\ref{eq:km-identity} holds and the $\update$ operator satisfies postulates U1-U4 and U8, then the $\erase$ operator satisfies postulates E1-E5 and E8.  The following counterexample demonstrates that Equation~\ref{eq:km-identity} does not hold on our operators: $P = \set{\belief{i}p}$ and $Q = \set{\possible{i}p, \possible{i}\neg p}$. From this, $P \erase Q = \set{}$, while $P \update \negkb{Q} = \set{\possible{i}p, \possible{i}\neg p}$, which when intersected with $\close{P}$ leaves $\set{\possible{i}p}$.

\citeasnoun{KM91} define a second identity between update and erasure:
\begin{align}
\label{eq:km-identity-two}
P \update Q &\equiv (P \erase \negkb{Q}) \conjoin Q
\end{align}

This  mirrors the Levi identity for belief revision and contraction \cite{levi1978subjunctives}.
They show that if  this identify holds and the $\erase$ operator satisfies E1-E4 and E8, then $\update$ satisfies U1-U4 and U8.  Equation~\ref{eq:km-identity-two} is just our definition of belief update, and \citeauthor{KM91}'s theorem about the relationship between the two sets of postulates holds:

\begin{theorem}
KM postulates E1, E3, and E4 hold for the \kb belief erasure operator. E2 holds if $P$ is satisfiable.
\end{theorem}

\begin{proof}
The proofs for E1, E3 and E4 are straightforward, so are omitted.

E2: Assume that the postulate does not hold. Then, there is some $\phi$ such that $P \erase Q \not\models \phi$ but $P \models \phi$ (the reverse cannot hold because $\erase$ is defined as set complement).  From the definitions of $\erase$ and downwards closure, this implies that $\close{P} \setminus \negkb{\close{\negkb{Q}}} \not\models \phi$. For this to hold, it must be that $\negkb{\close{\negkb{Q}}} \models \phi$, which implies that ${\negkb{Q}} \models \neg\phi$. However, from the premise $P \models \negkb{Q}$, this implies that $P \models \neg\phi$, which contradicts the assumption $P \models \phi$, so postulate E2 holds if $P$ is satisfiable.

If $P$ is unsatisfiable, postulate E2 not hold. A counterexample is $P = \set{p, \neg p}$ and $Q = \set{p}$.
\end{proof}

The E5 postulate does not hold. As a simple counterexample to this, consider the \kbs $P = \set{\belief{i}p}$ and $Q = \set{\possible{i}p}$. Erasing $Q$ from $P$ will result in an empty set, and then adding $Q$ will result in $\set{\possible{i}p}$, which does not entail $\belief{i}p$.

Finally, we note briefly on a controversial\footnote{See a discussion of the issues surrounding the postulate in Makinson \cite{makinson1987status}.} postulate --- the \emph{recovery postulate}:
\begin{align*}
(P \erase Q) \update Q &\models P
\end{align*}
This extends postulate E5, using $\update$ instead of $\conjoin$. The counterexample for E5 serves to show this postulate does not hold. However, we can characterise precisely when postulate E5 and the recovery postulate are satisfied: when  $Q = \dclose{Q}$. This follows directly from the definition of $\erase$.

\end{document}